\documentclass{article}

\usepackage[eandd,preprint,nonatbib]{neurips_2026}
\usepackage[numbers,compress]{natbib}

\usepackage[utf8]{inputenc}
\usepackage[T1]{fontenc}
\usepackage{xcolor}
\usepackage{colortbl}
\definecolor{linkred}{rgb}{0.87,0.44,0.38}
\definecolor{citeblue}{HTML}{699CDB}
\definecolor{scoregray}{gray}{0.92}
\definecolor{plotblue}{HTML}{1565C0}
\definecolor{plotred}{HTML}{C62828}
\definecolor{plotorange}{HTML}{F28E2B}
\definecolor{plotgreen}{HTML}{00897B}
\definecolor{plotpurple}{HTML}{8E24AA}
\usepackage[colorlinks=true,linkcolor=linkred,urlcolor=linkred,citecolor=citeblue]{hyperref}
\usepackage{url}
\usepackage{booktabs}
\usepackage{threeparttable}
\usepackage{multirow}
\usepackage{amsmath}
\usepackage{amssymb}
\usepackage{amsfonts}
\usepackage{graphicx}
\usepackage{subcaption}
\usepackage{wrapfig}
\usepackage[table-column-type=Z]{siunitx}
\usepackage{microtype}
\AtBeginDocument{}

\makeatletter
\renewcommand\paragraph{\@startsection{paragraph}{4}{\z@}%
  {0.5ex \@plus 0.2ex \@minus 0.1ex}%
  {-0.25em}%
  {\normalfont\normalsize\bfseries}}
\makeatother

\usepackage{tikz}
\usetikzlibrary{shapes.geometric}

\usepackage{pgfplots}
\usepgfplotslibrary{fillbetween,groupplots}
\pgfplotsset{compat=1.18}

\title{Driving on Memory}
\author{
Christian Löwens$^{1,3}$ \quad
Thorben Funke$^{1}$ \quad
Alexandru Paul Condurache$^{2,3}$
\vspace{0.2cm}
\\
$^{1}$Bosch Research \quad
$^{2}$Automated Driving, Bosch \quad
$^{3}$University of Lübeck
\vspace{0.2cm}
\\
\texttt{\{christian.loewens, thorben.funke, alexandrupaul.condurache\}@bosch.com}
}

\newcommand{\scorecell}[1]{#1}
\newcommand{\scorehead}[2]{\raisebox{-0.9ex}[3.6ex][1.8ex]{\shortstack{#1\\#2}}}
\newcommand{\refrowcell}[1]{{\hypersetup{citecolor=gray}\color{gray}\itshape #1}}
\newcommand{\refnumcell}[2]{\multicolumn{1}{>{\color{gray}\itshape}Z[table-format=#1]}{#2}}
\newcolumntype{S}{>{\columncolor{scoregray}}c}
\newcolumntype{E}{>{\columncolor{scoregray}[\tabcolsep][0pt]}c<{\hspace{\tabcolsep}}}
\newcolumntype{N}[1]{Z[table-format=#1]}
\newcolumntype{G}[1]{>{\columncolor{scoregray}}Z[table-format=#1,retain-explicit-plus=true]}
\newcolumntype{M}[1]{>{\columncolor{scoregray}[\tabcolsep][0pt]\centering\arraybackslash}m{#1}}
\newcolumntype{Q}[1]{>{\columncolor{scoregray}[\tabcolsep][\tabcolsep]\centering\arraybackslash}m{#1}}
\newenvironment{scoretabular}[1]{%
\begingroup
\setlength{\extrarowheight}{0.35ex}%
\setlength{\heavyrulewidth}{0.85pt}%
\setlength{\lightrulewidth}{0.65pt}%
\setlength{\cmidrulewidth}{0.45pt}%
\setlength{\aboverulesep}{0.30pt}%
\setlength{\belowrulesep}{0.30pt}%
\begin{tabular}{#1}%
}{%
\end{tabular}%
\endgroup
}
\newcommand{\cmark}{\ensuremath{\checkmark}}
\newcommand{\xmark}{\ensuremath{\times}}
\newcommand{\navtrain}{\texorpdfstring{\texttt{navtrain}}{navtrain}}
\newcommand{\navtest}{\texorpdfstring{\texttt{navtest}}{navtest}}
\newcommand{\navval}{\texorpdfstring{\texttt{navval}}{navval}}

\newcommand{\navhardtwostage}{\texorpdfstring{\texttt{navhard\_two\_stage}}{navhard-two-stage}}

\begin{document}

\maketitle


\begin{abstract}
End-to-end autonomous driving models plan future trajectories from raw sensor input. While earlier driving benchmarks often measured deviation from the human trajectory, current benchmarks such as NAVSIM and Bench2Drive evaluate models with richer simulation-based metrics intended to capture safe and compliant driving. A high benchmark score should reflect that a model can understand the scene in front of it and act accordingly. But how much of that score specifically comes from reacting to the dynamic part of that scene?

To probe this, we remove a model's camera input and replace it with memories from prior drives at the same location. The retrieved memories can provide persistent scene information, including road layout and location-conditioned regularities, but not the current traffic state. Surprisingly, memory is nearly sufficient on NAVSIM, reaching or even exceeding the performance of leading end-to-end methods without actually observing the evaluated scene. Our results suggest that a high NAVSIM score does not require a planner to react to the current traffic scene and should be treated with caution. This effect is benchmark-dependent: driving from memory causes substantially larger performance drops on Bench2Drive and RealEngine.
We provide our code at \href{https://github.com/boschresearch/MemoryDrivoR}{\nolinkurl{github.com/boschresearch/MemoryDrivoR}}.
\end{abstract}


\section{Introduction}
\label{sec:introduction}
End-to-end autonomous driving benchmarks can be read as tests of whether models understand the current driving situation. If a planner achieves a high benchmark score, the natural interpretation is that it has perceived the road layout, detected relevant traffic participants, inferred their intents and the rules of the scene, and planned accordingly. This interpretation is important: benchmark scores guide model design and leaderboard claims, and they are used to argue that modern driving systems are becoming better at scene understanding. Yet a high score does not by itself identify which information the model used. A benchmark can reward the intended capability, but it can also reward exploiting simpler cues that are correlated with the desired behavior.

This ambiguity has already appeared in end-to-end driving, where earlier benchmarks measured deviation from the human trajectory. It became especially visible on nuScenes, where a model without camera input can achieve surprisingly high benchmark scores from ego status alone, including speed and navigation command~\citep{li2024egostatus,zhai2023rethinking}. It goes without saying that this result does not show that perception is unnecessary for driving. It shows that high scores can come from exploiting regularities in the test rather than from having a holistic scene understanding. Recent benchmarks such as NAVSIM~\citep{cao2025pseudosimulation,dauner2024navsim} and Bench2Drive~\citep{jia2024bench2drive} were designed to include more demanding maneuvers, with simulation-based evaluation and richer metrics. Indeed, simple ego-status baselines perform much worse on these benchmarks. This motivates our question: do stronger benchmarks require models to understand the current situation, or can high scores still be achieved from comparatively weaker signals?

To this end, we distinguish between static and dynamic environmental information contained in the camera's input. Static information includes mostly map-like structure, but also persistent location priors such as typical crowdedness. Dynamic information includes the actual traffic participants, signal phase, weather, and other transient factors. In this work, we ask how much benchmark performance can be achieved from the static parts of the traffic scene alone.

\begin{figure}[t]
\centering
\includegraphics[width=\textwidth,trim=0 55 138 0,clip]{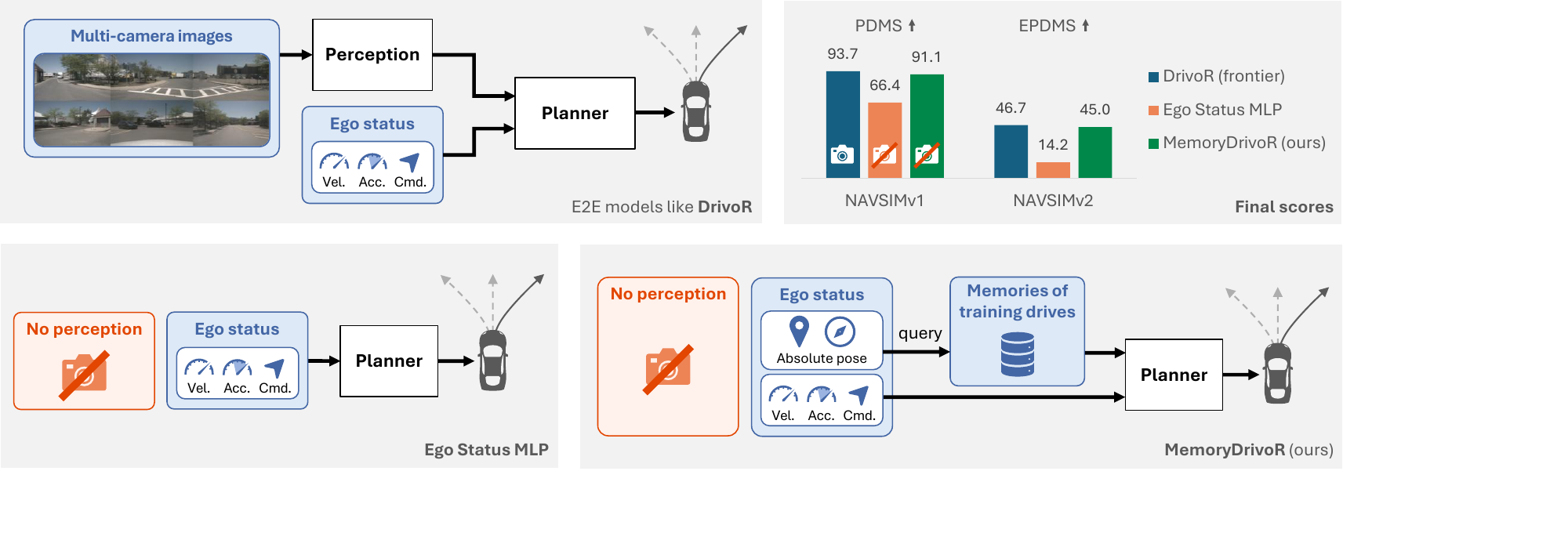}
\caption{\textbf{MemoryDrivoR.}
We introduce a novel analytic baseline for testing how much benchmark performance depends on camera input from the evaluated scene. MemoryDrivoR replaces camera input with memories from previous sensor-rich traversals. These memories carry static information but cannot describe the dynamics of the current traffic scene. In contrast to other baselines without cameras such as Ego Status MLP~\citep{li2024egostatus}, MemoryDrivoR achieves surprisingly strong NAVSIM performance and approaches leading camera-based methods like DrivoR~\citep{kirby2026drivor}, suggesting that the benchmark does not strongly demand reactions to the current situation.}
\label{fig:fig1}
\end{figure}

Thus, we introduce MemoryDrivoR, a new approach to instantiate this comparison as illustrated in Fig.~\ref{fig:fig1}. Starting from DrivoR~\citep{kirby2026drivor}, a transformer-based end-to-end planner, we build a memory bank from previous camera-based traversals in the training data. Each memory stores latent scene tokens together with the global pose at which they were observed. During inference, MemoryDrivoR retrieves nearby memories and injects them into the planner to replace the camera input of the evaluated scene. Thus, by design the planner can only access static and quasi-static information.

We apply this method to audit popular end-to-end benchmarks and find surprising results: MemoryDrivoR nearly reaches the performance of state-of-the-art camera-based models on both NAVSIMv1 and NAVSIMv2 by using this static information with additional cues from the initial ego status.
On Bench2Drive and RealEngine, the same intervention is not nearly as effective. We attribute this to their closed-loop evaluation with more demanding scenarios and, in the case of Bench2Drive, longer routes and an initial ego status that is not informative.
We also analyze camera-based performance under geographical splits to inspect whether models perceive static information at test time or memorize it during training. Our core contributions are:
\begin{itemize}
\item We introduce a novel stress test for end-to-end driving benchmarks by removing the camera input of the evaluated scene and substituting it with memories from previous drives.

\item Using our test, we show that NAVSIM scores can be very high even without seeing the current traffic scene. Our approach is close to state-of-the-art camera-based methods.

\item We show that other benchmarks, such as Bench2Drive and RealEngine, are robust to our baseline and discuss the resulting insights for the design of future benchmarks.

\item We distinguish our findings from location overfitting, which is not supported by the behavior of end-to-end models under geographical splits.
\end{itemize}


\section{Related Work}

\paragraph{Shortcut learning in end-to-end driving benchmarks.}
Several recent works caution that strong benchmark scores can arise from shortcuts that do not require perceiving the current traffic scene. AD-MLP~\citep{zhai2023rethinking} first exposed this issue on nuScenes~\citep{caesar2020nuscenes} open-loop planning, where predicted trajectories are compared with recorded futures. It showed that a simple non-perceptual planner using ego status, including navigation command, and past ego trajectory could achieve competitive scores. Ego-MLP~\citep{li2024egostatus} sharpened the critique by removing the reliance on past trajectories. It also showed that nuScenes contains many straightforward driving cases and introduced road-compliance analysis to reveal failures that displacement metrics alone hide.
The same concern also appears under richer planning scores. In nuPlan~\citep{karnchanachari2024nuplan}, PDM-Open~\citep{dauner2023parting} showed that a planner can achieve a high open-loop score by following a selected centerline while largely ignoring dynamic agents. We ask a related question: how far can benchmark performance be pushed when given additional static information? Instead of relying only on centerlines, we use previous drives as a richer source of persistent context.

\paragraph{Modern benchmark responses.}
Modern driving benchmarks explicitly try to avoid the weaknesses of open-loop trajectory matching. WOD-E2E~\citep{xu2026wode2e} is a large-scale end-to-end benchmark built on the Waymo Open Dataset~\citep{sun2020waymo}, with an emphasis on challenging long-tail scenarios. NAVSIMv1~\citep{dauner2024navsim} targets the ego status shortcut by filtering the OpenScene dataset~\citep{openscene2023} with a constant-velocity baseline to remove near-trivial scenes and replacing the displacement metric with a short non-reactive simulation scoring compliance, progress, and comfort. The paper interprets its ego-status MLP result as evidence that ego status remains useful but no longer competitive with sensor-based planners.

NAVSIMv2~\citep{cao2025pseudosimulation} further hardens the setup by adding a pseudo-simulation stage: the agent is evaluated not only from the recorded observation, but also from pre-generated synthetic viewpoints, while background traffic is made reactive. Other photorealistic simulators like HUGSIM~\citep{zhou2026hugsim} or RealEngine~\citep{jiang2025realengine} evaluate fully in closed loop, meaning the planner's actions affect subsequent simulator states. CARLA~\citep{dosovitskiy2017carla} is a common game-engine simulator for closed-loop driving~\citep{carlaLeaderboard20Evaluation,chitta2023transfuser,prakash2021multimodal} and is used by Bench2Drive~\citep{jia2024bench2drive} to design a benchmark with 44 interactive scenarios. Its authors report that AD-MLP fails on Bench2Drive, unlike on nuScenes, because the data distribution is less dominated by straight-driving cases. We use these benchmarks to ask whether shortcuts remain possible when additional static information is available.

\paragraph{Memory and map priors.}
MemoryDrivoR is related to a growing line of work that reuses information from previous drives. Neural Map Prior~\citep{xiong2023neural} builds and updates a learned global map prior from repeated traversals, then uses it to improve online mapping under occlusion and adverse conditions. PreSight~\citep{yuan2024presight} constructs a city-scale NeRF and reuses it to improve mapping and occupancy prediction. Others~\citep{li2024memorize,you2022hindsight,zhou2026compressed} incorporate memory to enhance background recognition in object detection. \citet{jia2026spatial} leverage Google Street View images as an additional input for perception, planning, and world-modeling. Concurrently, PriorEye~\citep{yeon2026prioreye} augments planners with route-conditioned memories derived from previous drives.

These works use historical context mainly as a performance aid for perception or planning. We use the same idea as an analytic probe: memory provides a proxy for static scene information, letting us test how much benchmark performance can be supported without observing the current traffic scene.


\section{Method}
\label{sec:method}
\begin{figure}
\centering
\includegraphics[width=\textwidth,trim=0 9 140 64,clip]{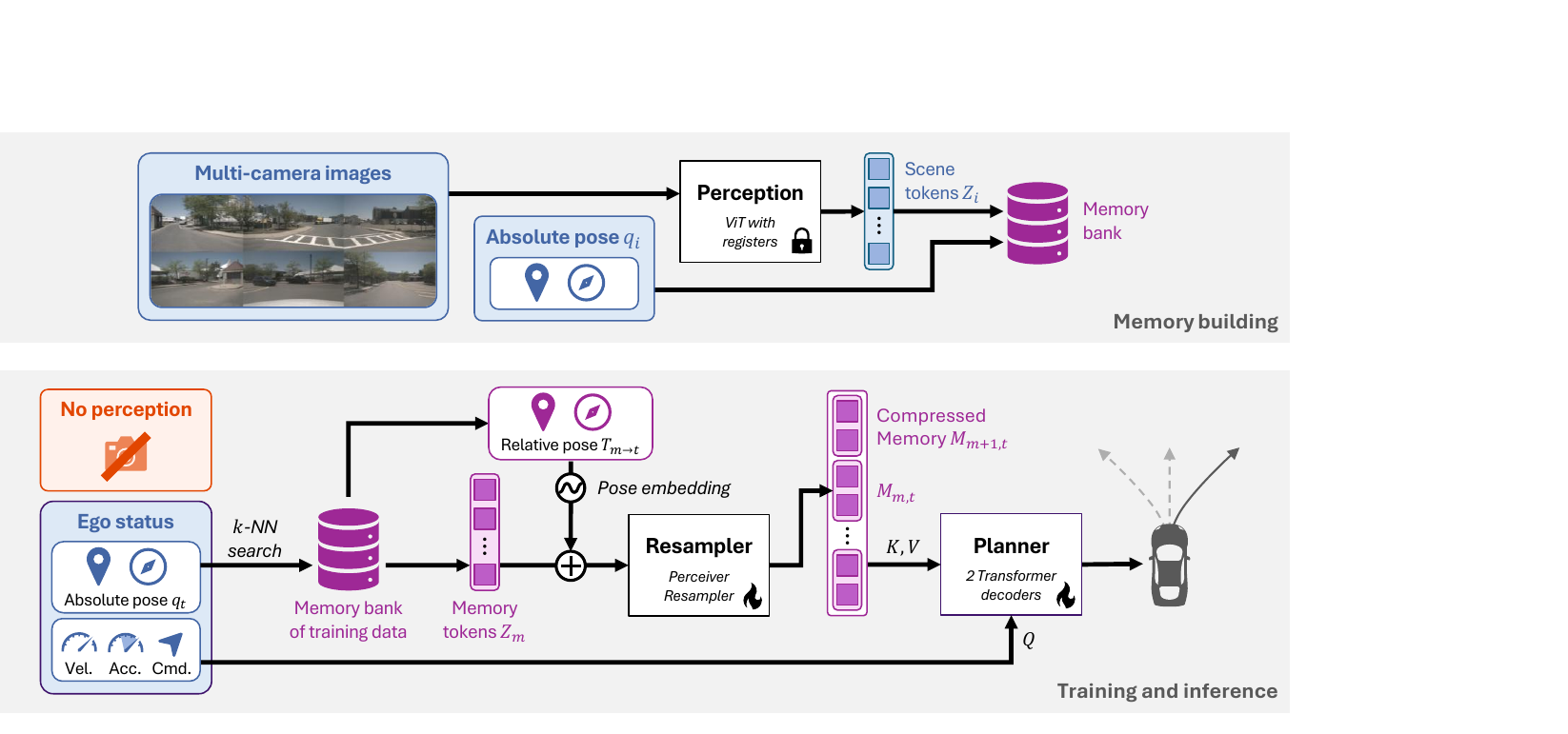}
\caption{\textbf{Method overview.} Previous camera-based traversals are encoded and stored in a pose-indexed memory bank. At query time, MemoryDrivoR removes online perception of the evaluated scene, retrieves nearby memories using the current pose, adds a relative-pose embedding, and passes the compressed memory tokens to the unchanged DrivoR planner.}
\label{fig:fig2}
\end{figure}

The key novelty of MemoryDrivoR is to audit end-to-end driving benchmarks by replacing camera observations with memories that approximate the scene's static information but provide no access to its current dynamics. We implement this intervention by storing previous camera-based traversals in a pose-indexed memory bank. At query time, the model cannot access images from the evaluated scene. Instead, it retrieves memories observed near the current position, conditions them on their relative pose, and uses them as context for the trajectory planner. Fig.~\ref{fig:fig2} summarizes our approach.

\subsection{Base Architecture}
We build on DrivoR~\citep{kirby2026drivor}, which uses a vision transformer (ViT) to encode multi-view camera images from a single frame into a set of register tokens~\citep{darcet2024vision} that summarize the current scene. Its planner module includes two transformer decoders: one proposes trajectories from ego-status queries, and the other scores those trajectories. Both decoders perform cross-attention with the scene tokens. The trajectory with the highest score is selected as the final prediction. We chose this architecture because its planner interface gives a controlled intervention point for compressed memory tokens.

\subsection{Memory Aggregation and Training}
\paragraph{Construction.} The memory bank is constructed from traversals of the training set.
For each selected frame \(i\), we run the frozen vision encoder from DrivoR and store the resulting scene tokens \(Z_i\in\mathbb{R}^{n\times c}\) with the ego pose \(q_i=(x_i,y_i,\psi_i)\), where \(x_i,y_i\) are ground-plane coordinates and \(\psi_i\) is heading. In what follows, we refer to the stored tokens as memory tokens.

\paragraph{Retrieval.} At query frame \(t\), MemoryDrivoR uses the current ego pose \(q_t\) to run a nearest-neighbor search over the stored bank poses. The retrieval operator returns up to \(k\) memories from eligible previous traversals whose stored positions lie within a radius \(r\), ranked by their distance. To ensure diversity, we select at most one memory from each traversal. We denote the returned indices by \(\mathcal{N}(t)\).

\paragraph{Injection.} Each retrieved memory is used as a latent representation from its original viewpoint, without geometric warping into the current camera view. Instead, we encode where the memory was observed relative to the current ego pose. For a pose \(q_i\), let \(T_i\in SE(2)\) be its homogeneous transform. For a memory pose \(q_m\) and current pose \(q_t\), we compute the relative transform as \(T_{m \rightarrow t}=T_t^{-1}T_m\).
A trainable pose embedder takes the flattened \(T_{m \rightarrow t}\), normalizes the translation by a fixed range, applies a sinusoidal encoding~\citep{mildenhall2020nerf}, and maps the result with a two-layer MLP to a \(c\)-dimensional embedding. The same pose embedding is then added to each memory token of \(Z_m\).

A learned transformer-based resampler~\citep{alayrac2022flamingo} compresses each pose-conditioned memory into a reduced set of \(\ell\) memory tokens, which are then projected by a linear layer to produce the final memory \(M_{m,t}\). The resampler learns to filter out outdated dynamic information about the environment that would otherwise distract the planner (see App.~\ref{app:resampler-probe}). We concatenate all compressed memories in $\mathcal{N}(t)$ and use them as a substitute for the keys $K$ and values $V$ of the two transformer decoders inside the planner. If no memories are available, we skip the corresponding cross-attention layers.
The downstream planner, including the trajectory decoder and scorer, is identical to the one from DrivoR.

\paragraph{Training.}
While the pose embedder and resampler are trained from scratch, we initialize the planner from pretrained DrivoR weights and fine-tune it on the memory tokens instead of camera-based tokens. MemoryDrivoR retains DrivoR's supervised objectives: an imitation loss over candidate trajectories and binary cross-entropy scorer losses for subscores such as no-at-fault collision, drivable-area and driving-direction compliance, time-to-collision, progress, and comfort.


\section{Experiments}
\label{sec:experiments}
We use MemoryDrivoR to audit two of the most widely adopted recent benchmarks in end-to-end driving~\citep{hu2025vlaad}, NAVSIM and Bench2Drive, and find markedly different demands for interaction. As supporting evaluations, we run the frozen NAVSIMv1 models on HUGSIM-nuScenes and RealEngine. We also ablate HD maps as an alternative to memory and additionally test for location overfitting.

\subsection{Main Benchmarks}
\paragraph{NAVSIMv1}\citep{dauner2024navsim} is a non-reactive simulation benchmark built on top of the OpenScene~\citep{openscene2023} redistribution of nuPlan~\citep{karnchanachari2024nuplan}, filtered to challenging scenarios. The agent is queried once on a real sensor frame and predicts a fixed \(4\,\mathrm{s}\) trajectory, which is then evaluated with background actors replaying their recorded futures. It reports the Predictive Driver Model Score (PDMS), which combines multiplicative penalties of no at-fault collisions (NC) and drivable area compliance (DAC) with a weighted average of time to collision (TTC), ego progress (EP), and comfort (C).

\paragraph{NAVSIMv2}\citep{cao2025pseudosimulation} extends this protocol with a pseudo-simulation stage: after the first rollout (Stage~1), the agent is re-evaluated on synthetic observations rendered around plausible future states with a 3D Gaussian Splatting~\citep{kerbl20233dgs} pipeline. In version 2, the surrounding traffic is reactive via the Intelligent Driver Model~\citep{treiber2000idm}. The headline metric is the extended PDM score (EPDMS), which adds lane keeping (LK), history comfort (HC), and extended comfort (EC) as weighted terms and driving direction compliance (DDC) and traffic light compliance (TLC) as multiplicative penalties.

\paragraph{Bench2Drive}\citep{jia2024bench2drive} is based on the CARLA~\citep{dosovitskiy2017carla} simulator.
It provides a training set collected by a privileged expert across \(12\) towns and \(23\) weather conditions.
The evaluation runs \(220\) routes with \(5\) routes per scenario. Unlike NAVSIM, it evaluates the agent fully in closed loop, with updated sensor input after each planning step. We report the two headline metrics, driving score (DS) and success rate (SR), along with the auxiliary subscores efficiency (Eff.) and comfort.

\subsection{Implementation Details}
\paragraph{Architecture.} The 2-layer resampler compresses the \(n=64\) register tokens from the frozen DrivoR ViT-S, conditioned on the relative pose, into \(\ell=8\) memory tokens. We ablate our pose embedder and resampler in App.~\ref{app:component-ablation}. The ego status is encoded into a single embedding and added to 64 learned trajectory queries. For NAVSIM, it concatenates the 2D ego velocity, 2D ego acceleration, and the 4D one-hot-encoded driving command. For Bench2Drive, we select only 1D speed and the 6D command.

\paragraph{Retrieval.} MemoryDrivoR retrieves up to \(k=10\) nearest memories within a radius of \(r=20\,\mathrm{m}\). As NAVSIM does not provide absolute ego poses in the official benchmark, we add them to enable pose-based retrieval. The NAVSIM memory bank contains all samples from the standard \navtrain{} split, while the Bench2Drive memory bank uses the larger Bench2Drive-Full training set to ensure coverage. App.~\ref{app:memory-retrieval-statistics} reports retrieval statistics and additional Bench2Drive filters.

\paragraph{Training.}
For NAVSIM, we initialize MemoryDrivoR with the DrivoR weights and fine-tune it for 5 additional epochs, keeping all other hyperparameters, including the learning rate and batch size, unchanged. For Bench2Drive, we fine-tune for 20 epochs starting from our DrivoR reimplementation, since the original model was not evaluated on this benchmark. Details are provided in App.~\ref{app:bench2drive-drivor}.

\paragraph{Baselines.} We compare MemoryDrivoR with the original camera-based DrivoR and with a no-camera DrivoR control, obtained by removing scene tokens from the planner. Concretely, we skip the cross-attention layers in both planner decoders, so trajectory proposal and scoring are conditioned only on the ego-status token. We also report leading camera-based end-to-end methods and simple baselines, including constant velocity and an ego status MLP.


\subsection{Results}
\label{sec:results}

\paragraph{NAVSIMv1.}
The final results for NAVSIMv1 are shown in Tab.~\ref{tab:navsim-v1-main}.
With camera input removed, DrivoR already improves over the ego-status MLP, so the architecture accounts for part of the gap.
When we add memory, MemoryDrivoR reaches a PDMS of 91.1, approaching the current camera-based state of the art.
The result suggests that most NAVSIMv1 benchmark performance can be achieved from ego status and static information, not from direct access to the current traffic scene.

\begin{figure*}[t]
    \centering
    \makebox[\textwidth][c]{\hspace*{-4.3pt}\definecolor{memonlyred}{HTML}{C62828}
\definecolor{egoonlypurple}{HTML}{8E24AA}
\definecolor{camerablue}{HTML}{0057FF}
\definecolor{memorypathgreen}{HTML}{00897B}
\definecolor{humangtorange}{HTML}{F28E2B}

\begin{tikzpicture}[
    legendline/.style={line width=1.15pt, dash pattern=on 2.8pt off 2.2pt},
    memorylegendline/.style={line width=1.15pt},
    legendtext/.style={anchor=west, font=\small, inner sep=0pt},
    x=1cm,
    y=1cm
]
    \coordinate (ego_start) at (0, 0);
    \draw[legendline, egoonlypurple] (ego_start) -- ++(0.56, 0);
    \draw[egoonlypurple, line width=1.0pt] (ego_start) ++(0.23, -0.07) -- ++(0.10, 0.14);
    \draw[egoonlypurple, line width=1.0pt] (ego_start) ++(0.23, 0.07) -- ++(0.10, -0.14);
    \path (ego_start) ++(0.68, 0) coordinate (ego_text_anchor);
    \node[legendtext] (ego_text) at (ego_text_anchor) {Ego};

    \path (ego_text.east) ++(0.35, 0) coordinate (camera_start);
    \draw[legendline, camerablue] (camera_start) -- ++(0.56, 0);
    \draw[camerablue, line width=1.0pt] (camera_start) ++(0.23, -0.08) -- ++(0.10, 0.16);
    \draw[camerablue, line width=1.0pt] (camera_start) ++(0.28, -0.10) -- ++(0, 0.20);
    \path (camera_start) ++(0.68, 0) coordinate (camera_text_anchor);
    \node[legendtext] (camera_text) at (camera_text_anchor) {Ego + camera};

    \path (camera_text.east) ++(0.35, 0) coordinate (mem_start);
    \draw[legendline, memonlyred] (mem_start) -- ++(0.56, 0);
    \path (mem_start) ++(0.28, 0) coordinate (mem_marker);
    \node[draw=memonlyred, fill=memonlyred, minimum size=3.5pt, inner sep=0pt] at (mem_marker) {};
    \path (mem_start) ++(0.68, 0) coordinate (mem_text_anchor);
    \node[legendtext] (mem_text) at (mem_text_anchor) {Ego + memory};

    \path (mem_text.east) ++(0.35, 0) coordinate (gt_start);
    \draw[legendline, humangtorange] (gt_start) -- ++(0.56, 0);
    \path (gt_start) ++(0.28, 0) coordinate (gt_marker);
    \node[diamond, draw=humangtorange, fill=humangtorange, minimum size=4.5pt, inner sep=0pt] at (gt_marker) {};
    \path (gt_start) ++(0.68, 0) coordinate (gt_text_anchor);
    \node[legendtext] (gt_text) at (gt_text_anchor) {Human};

    \path (gt_text.east) ++(0.35, 0) coordinate (memory_start);
    \draw[memorylegendline, memorypathgreen] (memory_start) -- ++(0.56, 0);
    \path (memory_start) ++(0.28, 0) coordinate (memory_marker);
    \node[circle, fill=memorypathgreen, minimum size=3.8pt, inner sep=0pt] at (memory_marker) {};
    \path (memory_start) ++(0.68, 0) coordinate (memory_text_anchor);
    \node[legendtext] at (memory_text_anchor) {Nearest memory path};
\end{tikzpicture}}
    \par
    \makebox[\textwidth][c]{%
        \begin{subfigure}[t]{0.386\textwidth}
            \centering
            \includegraphics[width=\linewidth]{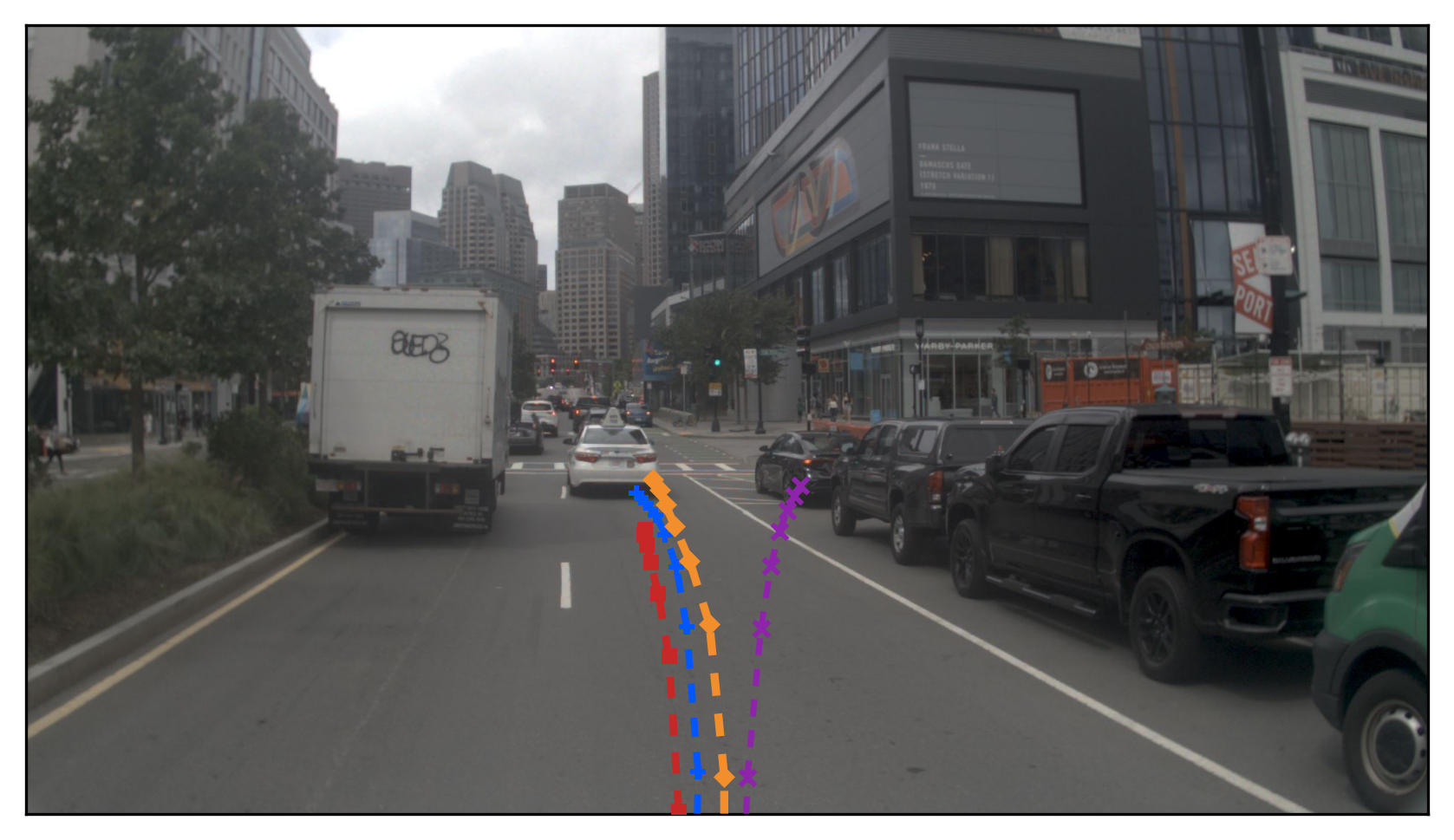}
            \caption{Current camera}
        \end{subfigure}%
        \hspace{0.012\textwidth}%
        \begin{subfigure}[t]{0.386\textwidth}
            \centering
            \includegraphics[width=\linewidth]{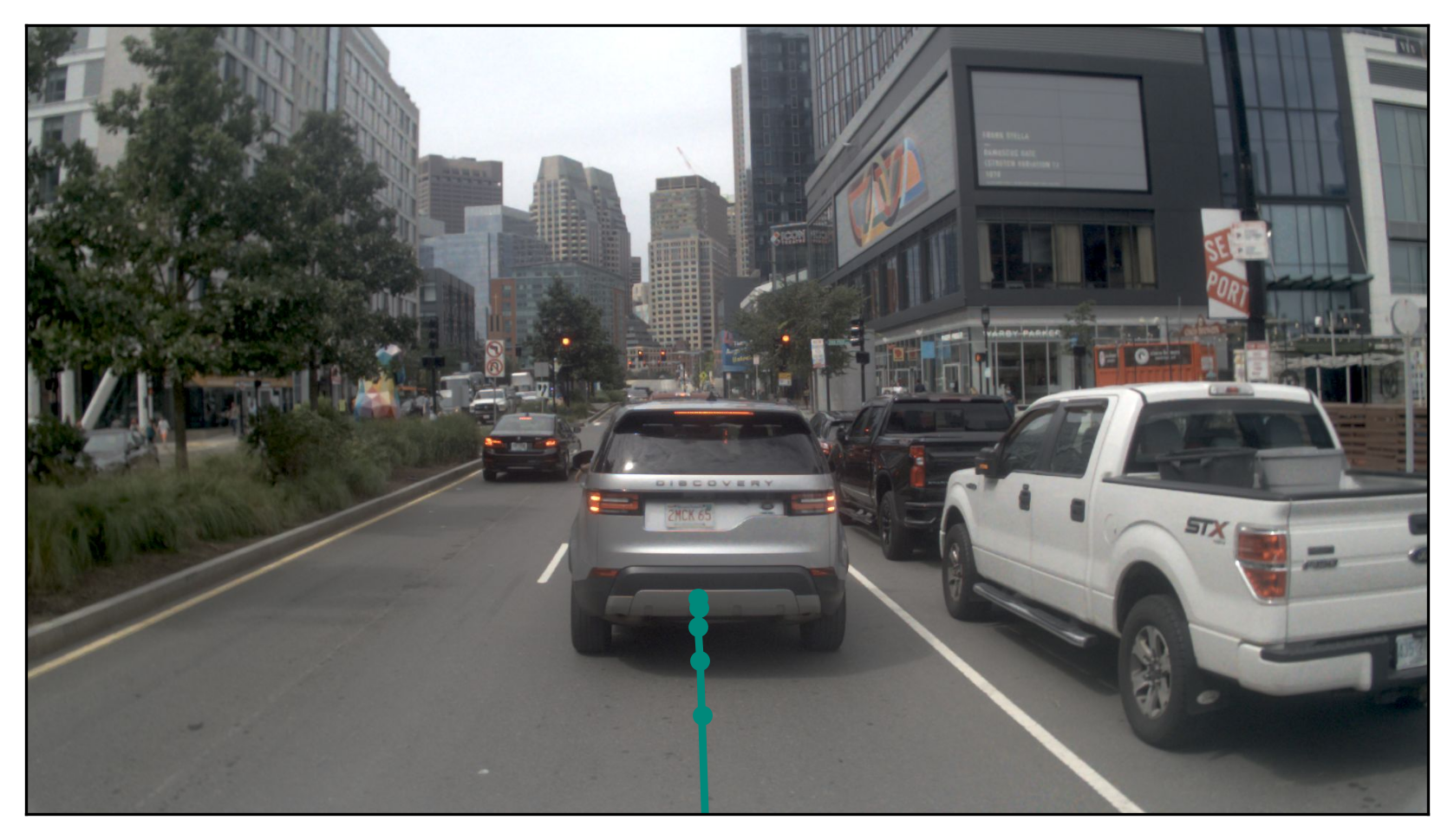}
            \caption{Retrieved memory}
        \end{subfigure}%
        \hspace{0.012\textwidth}%
        \begin{subfigure}[t]{0.204\textwidth}
            \centering
            \includegraphics[width=\linewidth]{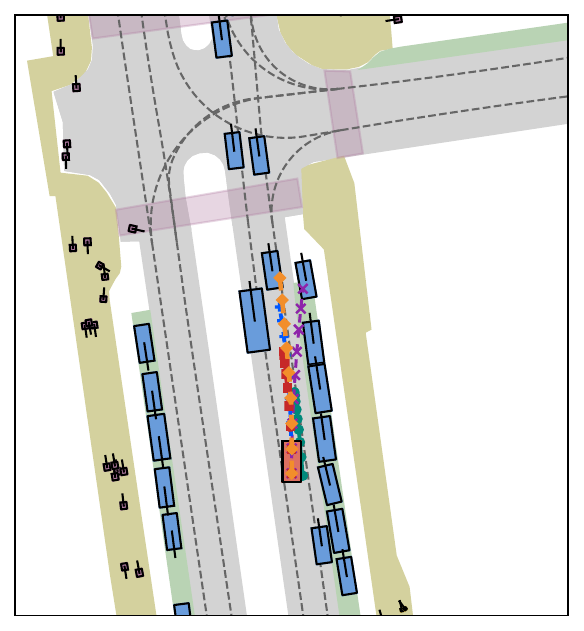}
            \caption{Paths}
        \end{subfigure}%
    }
    \caption{\textbf{Qualitative example.} We compare the planner prediction based on different sources of information on a NAVSIMv1 lane change. The center image is associated with the closest retrieved memory for the current sample. Each point marks a \SI{0.5}{\second} waypoint. More examples in App.~\ref{app:additional-qualitative-examples}.}
    \label{fig:mem_beats_ego_example}
\end{figure*}
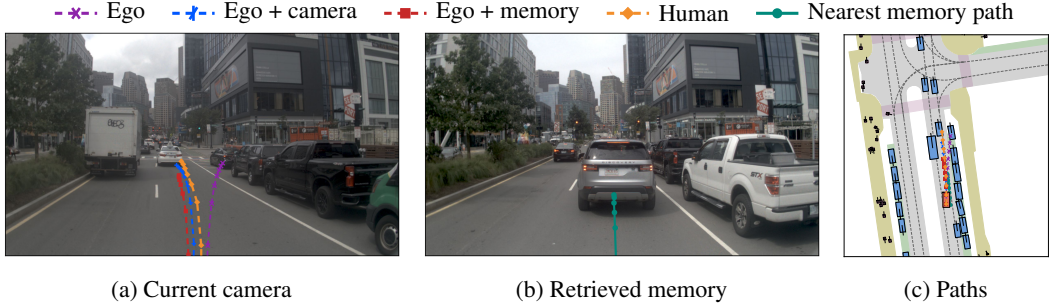

The DAC subscore changes least between DrivoR and MemoryDrivoR, consistent with memory preserving stable road geometry. However, even NC and TTC do not collapse despite their dependence on current actors.
As demonstrated in Fig.~\ref{fig:mem_beats_ego_example}, MemoryDrivoR drives generally more conservatively than DrivoR, especially at locations where dense traffic is likely. This helps to avoid collisions and reduce other noncompliant behavior as driving slower simply leads to fewer interactions over the evaluated \SI{4}{\second} horizon. Thus, the only substantial score reduction we measure is in ego progress.

We further investigated interesting scenes in which MemoryDrivoR stopped before a crosswalk or at a red traffic light. In most of those cases, the model already receives a braking signal from the initial ego status. Its memory of the static world, such as the crosswalk ahead, then provides the additional context needed to continue the braking. Since the evaluation horizon is short, the initial ego status is highly valuable to the model and already provides a cue about the anticipated maneuver.

The initial ego-status information, together with the defensive driving behavior, also explains the scenarios in which the model avoids collisions with a lead vehicle. In those cases, the ego-status baseline achieves similar performance even without memory, as static information is less important.

\begin{table}[t]
\centering
\setlength{\tabcolsep}{3pt}
\begin{threeparttable}
\caption{\textbf{NAVSIMv1} \navtest{} comparison with camera-based state-of-the-art methods.}
\label{tab:navsim-v1-main}
\begin{scoretabular}{llcccccccS}
\toprule
\multicolumn{1}{c}{Method} & \multicolumn{1}{c}{Venue} & Cam & Mem & NC & DAC & TTC & C & EP & \scorecell{PDMS~$\uparrow$} \\
\midrule
\refrowcell{PDM-Closed$^\ddagger$~\citep{dauner2023parting}} & \refrowcell{CoRL23} & \multicolumn{2}{c}{\refrowcell{privileged}} & \refrowcell{94.6} & \refrowcell{99.8} & \refrowcell{86.9} & \refrowcell{99.9} & \refrowcell{89.9} & \scorecell{\refrowcell{89.1}} \\
\refrowcell{Human driver~\citep{dauner2024navsim}} &  & \multicolumn{2}{c}{\refrowcell{-}} & \refrowcell{100.0} & \refrowcell{100.0} & \refrowcell{100.0} & \refrowcell{99.9} & \refrowcell{87.5} & \scorecell{\refrowcell{94.8}} \\
\midrule
UniAD$^\S$~\citep{hu2023planning} & CVPR23 & \cmark & \xmark & 97.8 & 91.9 & 92.9 & 100.0 & 78.8 & \scorecell{83.4} \\
LTFv6$^\dagger$~\citep{nguyen2026lead} & CVPR26 & \cmark & \xmark & 97.5 & 95.4 & 93.8 & 100.0 & 80.9 & \scorecell{86.4} \\
DiffusionDrive$^\dagger$~\citep{liao2025diffusiondrive} & CVPR25 & \cmark & \xmark & 98.2 & 96.2 & 94.7 & 100.0 & 82.2 & \scorecell{88.0} \\
AutoVLA~\citep{zhou2025autovla} & NeurIPS25 & \cmark & \xmark & 98.4 & 95.6 & 98.0 & 99.9 & 81.9 & \scorecell{89.1} \\
ReCogDrive~\citep{li2026recogdrive} & ICLR26 & \cmark & \xmark & 97.9 & 97.3 & 94.9 & 100.0 & 87.3 & \scorecell{90.8} \\
DriveSuprim~\citep{yao2026drivesuprim} & AAAI26 & \cmark & \xmark & 98.6 & 98.6 & 95.5 & 100.0 & 91.3 & \scorecell{93.5} \\
RAP-DINO$^\dagger$~\citep{feng2026rap} & ICLR26 & \cmark & \xmark & 99.1 & 98.9 & 96.7 & 100.0 & 90.3 & \scorecell{\textbf{93.8}} \\
DrivoR~\citep{kirby2026drivor} & CVPR26 & \cmark & \xmark & 99.0 & 98.9 & 96.7 & 100.0 & 90.0 & \scorecell{93.7} \\
\midrule
Constant Velocity$^\dagger$ &  & \xmark & \xmark & 68.0 & 57.8 & 50.0 & 100.0 & 19.4 & \scorecell{20.7} \\
Ego Status MLP$^\dagger$ &  & \xmark & \xmark & 93.1 & 78.3 & 84.0 & 100.0 & 63.2 & \scorecell{66.4} \\
DrivoR &  & \xmark & \xmark & 96.1 & 85.6 & 91.1 & 100.0 & 62.8 & \scorecell{72.9} \\
MemoryDrivoR (ours) &  & \xmark & \cmark & 98.3 & 98.5 & 95.3 & 100.0 & 86.0 & \scorecell{\textbf{91.1}} \\
\bottomrule
\end{scoretabular}
\begin{tablenotes}[flushleft]
\footnotesize
\item[] $^\dagger$ Results from the public leaderboard~\citep{agc2024navtestleaderboard}. $^\ddagger$ Results from~\citep{li2024hydra}. $^\S$ Results from~\citep{dauner2024navsim}.
\end{tablenotes}
\end{threeparttable}
\end{table}

\paragraph{NAVSIMv2}
 preserves the same qualitative story, as shown in Tab.~\ref{tab:navsim-v2-main}. Even without cameras, the stronger DrivoR architecture improves markedly over the ego-status MLP and approaches camera-based methods such as LTFv6~\citep{nguyen2026lead}, a result that has not previously been shown. Adding memory brings it close to DrivoR and well above other recent camera-based methods. This remains true even though NAVSIMv2 is more realistic than NAVSIMv1, with reactive traffic and shifted Stage~2 starting points.
The component scores are close to those of DrivoR, with slightly stronger results on metrics tied to static information. Again, the metrics most tied to dynamic information, such as collision and traffic-light compliance, do not substantially drop. The main tradeoff appears in ego progress.

\begin{table}[t]
\centering
\small
\setlength{\tabcolsep}{2pt}
\begin{threeparttable}
\caption{\textbf{NAVSIMv2} \navhardtwostage{} comparison with camera-based methods.}
\label{tab:navsim-v2-main}
\begin{scoretabular}{>{\raggedright\arraybackslash}p{2.75cm}ccccccccccccS}
\toprule
\multicolumn{1}{c}{Method} & Cam & Mem & Stage & NC & DAC & DDC & TLC & EP & TTC & LK & HC & EC & \scorecell{EPDMS~$\uparrow$} \\
\midrule
\multirow{2}{*}{\refrowcell{PDM-Closed$^\ddagger$~\citep{dauner2023parting}}}
 & \multicolumn{2}{c}{\multirow{2}{*}{\refrowcell{privileged}}} & \refrowcell{S1} & \refrowcell{94.4} & \refrowcell{98.8} & \refrowcell{100.0} & \refrowcell{99.5} & \refrowcell{100.0} & \refrowcell{93.5} & \refrowcell{99.3} & \refrowcell{87.7} & \refrowcell{36.0} & \\
 & \multicolumn{2}{c}{} & \refrowcell{S2} & \refrowcell{90.5} & \refrowcell{90.6} & \refrowcell{95.4} & \refrowcell{98.4} & \refrowcell{100.0} & \refrowcell{86.6} & \refrowcell{74.2} & \refrowcell{91.9} & \refrowcell{29.7} & \scorecell{\multirow{-2}{*}{\refrowcell{56.6}}} \\
\midrule
\multirow{2}{*}{LTFv6$^\dagger$~\citep{nguyen2026lead}} & \multirow{2}{*}{\cmark} & \multirow{2}{*}{\xmark} & S1 & 96.6 & 86.7 & 99.2 & 99.6 & 84.5 & 95.1 & 94.4 & 97.8 & 76.4 & \\
 &  &  & S2 & 79.9 & 75.6 & 86.3 & 97.9 & 89.6 & 76.1 & 50.1 & 95.2 & 66.7 & \scorecell{\multirow{-2}{*}{31.9}} \\
\multirow{2}{*}{RAP-DINO$^\dagger$~\citep{feng2026rap}} & \multirow{2}{*}{\cmark} & \multirow{2}{*}{\xmark} & S1 & 97.1 & 94.4 & 98.8 & 99.8 & 83.9 & 96.9 & 94.7 & 96.4 & 66.2 & \\
 &  &  & S2 & 83.2 & 83.9 & 87.4 & 98.0 & 86.9 & 80.4 & 52.3 & 95.2 & 52.4 & \scorecell{\multirow{-2}{*}{39.6}} \\
\multirow{2}{*}{ZTRS$^\dagger$~\citep{li2025ztrs}} & \multirow{2}{*}{\cmark} & \multirow{2}{*}{\xmark} & S1 & 98.9 & 97.6 & 100.0 & 100.0 & 66.7 & 98.9 & 96.2 & 96.7 & 44.0 & \\
 &  &  & S2 & 91.1 & 90.4 & 95.8 & 99.0 & 63.6 & 89.8 & 60.4 & 97.6 & 66.1 & \scorecell{\multirow{-2}{*}{\textbf{48.1}}} \\
\multirow{2}{*}{DrivoR$^\ast$~\citep{kirby2026drivor}} & \multirow{2}{*}{\cmark} & \multirow{2}{*}{\xmark} & S1 & 99.3 & 95.8 & 99.3 & 99.8 & 73.5 & 99.3 & 94.2 & 97.6 & 70.7 & \\
 &  &  & S2 & 89.8 & 86.4 & 90.8 & 98.3 & 71.3 & 88.2 & 51.8 & 99.0 & 76.8 & \scorecell{\multirow{-2}{*}{46.7}} \\
\midrule
\multirow{2}{*}{Constant Velocity$^\dagger$} & \multirow{2}{*}{\xmark} & \multirow{2}{*}{\xmark} & S1 & 88.9 & 42.9 & 70.7 & 99.3 & 77.5 & 87.3 & 78.7 & 97.1 & 60.4 & \\
 &  &  & S2 & 83.2 & 59.1 & 76.5 & 98.1 & 71.4 & 81.1 & 48.0 & 97.2 & 62.0 & \scorecell{\multirow{-2}{*}{11.5}} \\
\multirow{2}{*}{Ego Status MLP$^\dagger$} & \multirow{2}{*}{\xmark} & \multirow{2}{*}{\xmark} & S1 & 93.2 & 55.8 & 86.7 & 99.3 & 81.2 & 92.2 & 83.6 & 97.6 & 77.8 & \\
 &  &  & S2 & 77.2 & 51.9 & 74.4 & 98.3 & 77.1 & 75.1 & 40.9 & 97.8 & 79.9 & \scorecell{\multirow{-2}{*}{14.2}} \\
\multirow{2}{*}{DrivoR} & \multirow{2}{*}{\xmark} & \multirow{2}{*}{\xmark} & S1 & 95.4 & 72.2 & 93.7 & 100.0 & 67.4 & 94.4 & 83.8 & 97.6 & 68.4 & \\
 &  &  & S2 & 88.2 & 71.7 & 87.8 & 99.3 & 56.1 & 86.1 & 48.1 & 99.0 & 80.9 & \scorecell{\multirow{-2}{*}{28.3}} \\
MemoryDrivoR & \multirow{2}{*}{\xmark} & \multirow{2}{*}{\cmark} & S1 & 98.2 & 95.8 & 99.0 & 100.0 & 72.0 & 98.0 & 92.0 & 97.6 & 69.3 & \\
(ours) &  &  & S2 & 88.2 & 89.1 & 92.2 & 98.7 & 58.4 & 87.6 & 52.0 & 98.3 & 75.2 & \scorecell{\multirow{-2}{*}{\textbf{45.0}}} \\
\bottomrule
\end{scoretabular}
\begin{tablenotes}[flushleft]
\footnotesize
\item[] $^\dagger$ Results from the public leaderboard~\citep{agc2025navhardleaderboard}. $^\ddagger$ Results from~\citep{cao2025pseudosimulation}. $^\ast$ Results reproduced by us.
\end{tablenotes}
\end{threeparttable}
\end{table}

\begin{wrapfigure}[12]{r}{0.355\textwidth}
    \centering
    \captionsetup{font=footnotesize,skip=2pt}
        \includegraphics[width=\linewidth]{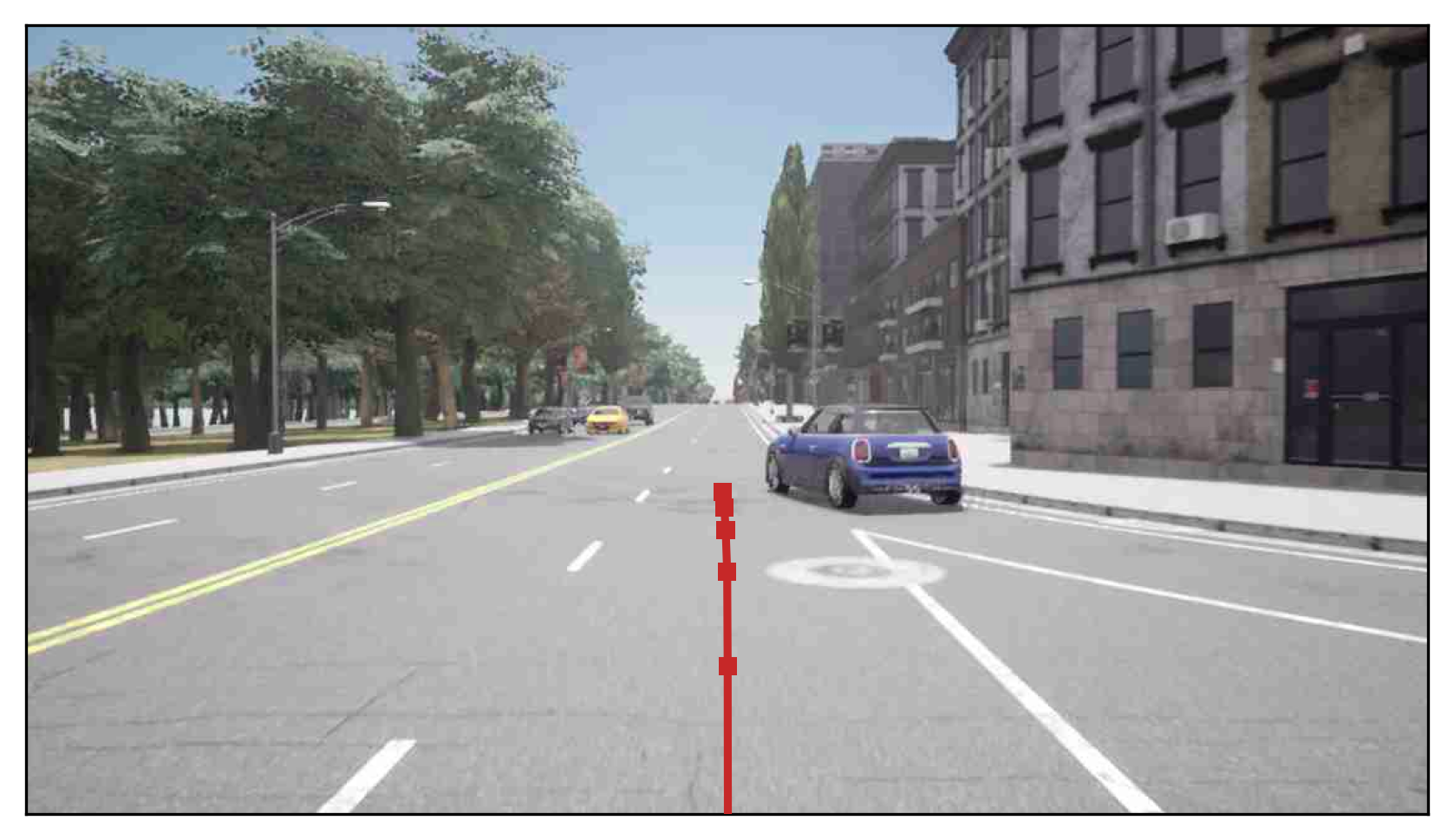}
    \caption{\textbf{Successes in Bench2Drive.} In less complex HighwayCutIn scenarios, MemoryDrivoR learns to slow down at the merge. Waypoints are shown at \SI{0.5}{\second} intervals.}
    \label{fig:b2d-same-route}
\end{wrapfigure}
\paragraph{Bench2Drive.}
The Bench2Drive results are shown in Tab.~\ref{tab:b2d-main}, with the corresponding multi-ability breakdown and failure analysis in App.~\ref{app:failure-analysis}.
MemoryDrivoR substantially outperforms the no-camera baselines but remains far below the camera-based model and stronger planners.
The routes successfully completed by MemoryDrivoR average about $\SI{110}{\meter}$ and $\SI{79}{\second}$. While still short compared to other CARLA benchmarks~\citep{prakash2021multimodal,chitta2023transfuser,carlaLeaderboard20Evaluation}, this is far beyond the $\SI{4}{\second}$ horizon evaluated in NAVSIMv1 and makes it more challenging to drive from memory alone. We evaluate collisions by distance traveled in App.~\ref{app:controlled-horizon} and find that the performance gap between MemoryDrivoR and DrivoR widens as travel distance increases. In contrast to NAVSIM, the initial ego status does not provide any meaningful information that could be exploited since each scenario starts with the vehicle standing still.

However, MemoryDrivoR still completes 21 of 220 routes without infractions.
The most striking scenario type is the HighwayCutIn shown in Fig.~\ref{fig:b2d-same-route}, where the planner completes all five routes.
Although another vehicle ahead merges from an on-ramp, the scene contains very few other agents overall and can be handled by always slowing down before the merge.

\begin{table}[t]
\centering
\setlength{\tabcolsep}{3pt}
\begin{threeparttable}
\caption{\textbf{Bench2Drive} closed-loop comparison with camera-based methods.}
\label{tab:b2d-main}
\begin{scoretabular}{llccN{3.1}N{2.1}M{0.88cm}@{}Q{0.88cm}}
\toprule
\multicolumn{1}{c}{Method} & \multicolumn{1}{c}{Venue} & Cam & Mem & \multicolumn{1}{c}{Eff.} & \multicolumn{1}{c}{Comfort} & \scorecell{SR~$\uparrow$} & \scorecell{DS~$\uparrow$} \\
\midrule
\refrowcell{Think2Drive$^\dagger$~\citep{li2024think2drive}} & \refrowcell{ECCV24} & \multicolumn{2}{c}{\refrowcell{privileged}} & \refnumcell{3.1}{269.1} & \refnumcell{2.1}{26.0} & \refrowcell{\tablenum[table-format=2.1]{85.4}} & \refrowcell{\tablenum[table-format=2.1]{91.9}} \\
\refrowcell{LEAD~\citep{nguyen2026lead}} & \refrowcell{CVPR26} & \multicolumn{2}{c}{\refrowcell{privileged}} & \multicolumn{1}{c}{\refrowcell{--}} & \multicolumn{1}{c}{\refrowcell{--}} & \refrowcell{\tablenum[table-format=2.1]{92.3}} & \refrowcell{\tablenum[table-format=2.1]{97.0}} \\
\midrule
UniAD$^\dagger$~\citep{hu2023planning} & CVPR23 & \cmark & \xmark & 129.2 & 43.6 & \tablenum[table-format=2.1]{16.4} & \tablenum[table-format=2.1]{45.8} \\
RAP-ResNet~\citep{feng2026rap} & ICLR26 & \cmark & \xmark & 165.5 & 23.6 & \tablenum[table-format=2.1]{37.3} & \tablenum[table-format=2.1]{66.4} \\
ReCogDrive~\citep{li2026recogdrive} & ICLR26 & \cmark & \xmark & 138.2 & 17.5 & \tablenum[table-format=2.1]{45.5} & \tablenum[table-format=2.1]{71.4} \\
AutoVLA~\citep{zhou2025autovla} & NeurIPS25 & \cmark & \xmark & 146.9 & 39.3 & \tablenum[table-format=2.1]{57.7} & \tablenum[table-format=2.1]{78.8} \\
DriveSuprim~\citep{yao2026drivesuprim} & AAAI26 & \cmark & \xmark & 238.8 & 20.9 & \tablenum[table-format=2.1]{60.0} & \tablenum[table-format=2.1]{83.0} \\
SimLingo~\citep{renz2025simlingo} & CVPR25 & \cmark & \xmark & 244.2 & 25.5 & \bfseries\tablenum[table-format=2.1]{66.8} & \bfseries\tablenum[table-format=2.1]{85.9} \\
DrivoR$^\ast$~\citep{kirby2026drivor} & CVPR26 & \cmark & \xmark & 167.8 & 36.5 & \tablenum[table-format=2.1]{30.5} & \tablenum[table-format=2.1]{61.0} \\
\midrule
AD-MLP$^\dagger$~\citep{zhai2023rethinking} & & \xmark & \xmark & 48.5 & 22.6 & \tablenum[table-format=2.1]{0.0} & \tablenum[table-format=2.1]{18.1} \\
DrivoR$^\ast$ &  & \xmark & \xmark & 225.3 & 86.9 & \tablenum[table-format=2.1]{2.4} & \tablenum[table-format=2.1]{11.6} \\
MemoryDrivoR (ours) &  & \xmark & \cmark & 176.4 & 49.9 & \bfseries\tablenum[table-format=2.1]{9.5} & \bfseries\tablenum[table-format=2.1]{34.7} \\
\bottomrule
\end{scoretabular}
\begin{tablenotes}[flushleft]
\footnotesize
\item[] $^\dagger$ Results from~\citep{yang2025raw2drive}. $^\ast$ Results are based on our own implementation.
\end{tablenotes}
\end{threeparttable}
\end{table}

\subsection{Additional Benchmarks}
Furthermore, we test our NAVSIMv1 models on two additional photorealistic closed-loop benchmarks, namely HUGSIM-nuScenes~\citep{zhou2026hugsim} and RealEngine~\citep{jiang2025realengine}.

\begin{table}[t]
\centering
\footnotesize
\setlength{\tabcolsep}{3pt}
\begin{threeparttable}
\captionsetup{hypcap=false}
\captionof{table}{\textbf{RealEngine PDMS} across evaluation protocols on NAVSIM scenes.}
\label{tab:experimental-realengine}
\begin{scoretabular}{lcccSSSS}
\toprule
Method & Cam & LiDAR & Mem & \scorecell{Open-Loop~$\uparrow$} & Base Closed-Loop~$\uparrow$ & Safety test~$\uparrow$ & \scorecell{Multi-agent~$\uparrow$} \\
\midrule
DiffusionDrive$^\dagger$~\citep{liao2025diffusiondrive} & \cmark & \cmark & \xmark & 69.5 & 61.3 & \textbf{53.8} & 51.9 \\
DrivoR & \cmark & \xmark & \xmark & \textbf{86.6} & \textbf{88.2} & \textbf{53.8} & \textbf{63.9} \\
\midrule
Constant Velocity$^\dagger$ & \xmark & \xmark & \xmark & 46.8 & 46.8 & 36.3 & 27.4\\
DrivoR & \xmark & \xmark & \xmark & 65.6 & 55.1 & \textbf{39.2} & \textbf{51.9} \\
MemoryDrivoR (ours) & \xmark & \xmark & \cmark & \textbf{87.1} & \textbf{72.3} & 28.1 & 43.4 \\
\bottomrule
\end{scoretabular}
\begin{tablenotes}[flushleft]
\footnotesize
\item[] $^\dagger$ Results from~\citep{jiang2025realengine}.
\end{tablenotes}
\end{threeparttable}
\end{table}

\paragraph{RealEngine} provides a non-reactive open- and closed-loop evaluation of 14 NAVSIMv1 scenes. In addition, it executes 21 safety tests with synthetically inserted counterfactual actors and 28 multi-agent interactions in which two instances of the evaluated planner model are deployed. The open-loop scores in Tab.~\ref{tab:experimental-realengine} are consistent with our results on NAVSIM. However, in a true closed-loop setting, the gap between DrivoR and both no-camera baselines grows substantially, even though the evaluation uses the same short horizon. After the first planning step in closed loop, the initial ground-truth ego status is not available to the models anymore, since this information gets updated at each timestamp, leading to uncharted state spaces. An architecture that stores the historic status could prevent this. Moreover, the gap for MemoryDrivoR increases when RealEngine inserts additional actors. The collision-related metrics account for most of this loss.

\setlength{\intextsep}{5pt}
\begin{wraptable}[10]{r}{0.34\textwidth}
\vspace{0.5\baselineskip}
\centering
\captionsetup{font=footnotesize,skip=2pt}
\scriptsize
\setlength{\tabcolsep}{1.25pt}
\begin{threeparttable}
\caption{\textbf{HUGSIM-nuScenes} 0-shot evaluation of the DrivoR models}
\label{tab:hugsim-nuscenes}
\begin{scoretabular}{@{}lccE@{}}
\toprule
\multicolumn{1}{c}{\makebox[1.5cm]{Method}} & \multicolumn{1}{c}{Cam} & \multicolumn{1}{c}{Mem} & \scorecell{HD-Score~$\uparrow$} \\
\midrule
UniAD$^\dagger$ & \cmark &\xmark & \bfseries \scorecell{36.7} \\
DrivoR & \cmark & \xmark & \scorecell{36.4} \\
\midrule
DrivoR & \xmark & \xmark & \scorecell{29.1} \\
MemoryDrivoR & \xmark & \cmark & \bfseries\scorecell{32.2} \\
\bottomrule
\end{scoretabular}
\begin{tablenotes}[flushleft]
\scriptsize
\item[] $^\dagger$ Trained on nuScenes.
\end{tablenotes}
\end{threeparttable}
\end{wraptable}
\paragraph{HUGSIM-nuScenes} evaluates closed-loop episodes, which last up to \SI{56}{\second} in case of MemoryDrivoR. As shown in Tab.~\ref{tab:hugsim-nuscenes}, MemoryDrivoR attains a HUGSIM driving score (HD-Score) of 32.2\%, compared with 36.4\% for the camera-based DrivoR. We do not observe a performance gap like in Bench2Drive or RealEngine. This may be because that the 87 scenarios in HUGSIM-nuScenes are derived from 12 mostly straight routes and therefore generally simple to solve. This result shows that MemoryDrivoR can compete with camera-based models even over longer evaluation horizons in closed loop when interaction with the dynamic world is limited. Further details are provided in App.~\ref{app:additional-benchmark-details}.

\setlength{\intextsep}{5pt}
\begin{wraptable}[8]{r}{0.405\textwidth}
\vspace{-0.5\baselineskip}
\centering
\captionsetup{font=footnotesize,skip=2pt}
\footnotesize
\setlength{\tabcolsep}{2pt}
\begin{threeparttable}
\caption{\textbf{HD-map} ablation for DrivoR inputs.}
\label{tab:hdmap-comparison}
\begin{scoretabular}{@{}lSE@{}}
\toprule
\multicolumn{1}{c}{\makebox[2cm]{Input}} & \scorecell{\scorehead{NAVSIMv1}{PDMS~$\uparrow$}} & \scorecell{\scorehead{NAVSIMv2}{EPDMS~$\uparrow$}} \\
\midrule
\refrowcell{Ego + cameras} & \scorecell{\refrowcell{93.7}} & \scorecell{\refrowcell{46.7}} \\
\midrule
Ego & \scorecell{72.9} & \scorecell{28.3} \\
Ego + HD map & \scorecell{86.6} & \scorecell{40.8} \\
Ego + memory & \scorecell{\textbf{91.1}} & \scorecell{\textbf{45.0}} \\
\bottomrule
\end{scoretabular}
\end{threeparttable}
\end{wraptable}

\subsection{Ablations}
\label{sec:ablations}
\paragraph{Would HD maps show the same effect?}
\label{sec:hd-maps}
A natural question is whether an HD map could provide the same audit signal, similar in spirit to a centerline-based planner such as PDM-Open~\citep{dauner2023parting}. We treat this as a narrower baseline: unlike an HD map, memory can also encode coarse location-conditioned regularities such as typical crowdedness or recurring congestion. These remain priors rather than observations of the current scene.

To examine the difference, we modify DrivoR to use HD-map inputs. We adapt the BERT-based vector encoder of PriorDrive~\citep{zeng2026priordrive} for nearby static map elements and their attributes. The resulting map tokens replace the camera-derived scene tokens used by the planner, analogously to our memory-token injection. For full implementation details and component results, we refer the reader to App.~\ref{app:hd-map}.

The NAVSIM results are shown in Tab.~\ref{tab:hdmap-comparison}.
The map-baseline closes a substantial part of the gap to camera-based DrivoR, but remains below MemoryDrivoR. These results support the hypothesis that memory can capture richer static information and is therefore a better source for this kind of audit.

\paragraph{Does memorization explain performance?}
The preceding results show that high NAVSIM performance can be achieved from static information. This makes it important to rule out a more prosaic explanation: that the model was rewarded for memorizing test locations during training, rather than perceiving static scene information at test time.
For NAVSIM and Bench2Drive, training and test overlap substantially. This leaves open the possibility that models may overfit to location-specific regularities already encoded in their weights, a failure mode documented in online mapping~\citep{yuan2024streammapnet,lilja2024localization}.

We therefore run evaluations on both benchmarks using geographical subsets. In NAVSIMv2-Geo, we remove Pittsburgh samples from training, retaining approximately \(80\%\) of the original training set, and evaluate only on Pittsburgh scenarios, which constitute \(40\%\) of the original test. In Bench2Drive-Geo, we remove Town11 and Town13 from training, again retaining \(80\%\) of the original trainval set, and evaluate only on those towns, which constitute \(25\%\) of the original test set.
The relevant comparison is therefore between \emph{camera-based} DrivoR trained on the standard split and the same model trained on the geographical split. If performance were primarily caused by memorization of those locations, removing them from training should substantially reduce performance.

As shown in Tab.~\ref{tab:geo-splits}, we observe no such degradation. After removing the test locations from the training data, the overall benchmark score even increased slightly in both cases. For the DrivoR models evaluated here, this makes memorization of geography an unlikely primary explanation.
Geographical splits therefore control for location memorization, but they do not fix the benchmark behavior exposed by our earlier experiments: a model evaluated on held-out geography can still perceive static information at test time.

\begin{table}[t]
\centering
\caption{\textbf{Geographical split} results for DrivoR on held-out evaluation locations.}
\label{tab:geo-splits}
\begin{subtable}[t]{0.5\textwidth}
\centering
\footnotesize
\setlength{\tabcolsep}{2pt}
\caption{NAVSIMv2-Geo (Pittsburgh)}
\label{tab:navsim-v2-geo}
\begin{scoretabular}{@{}lE@{}}
\toprule
\multicolumn{1}{@{}c}{Training locations} & EPDMS~$\uparrow$ \\
\midrule
Original (all cities) & 39.0 \\
w/o Pittsburgh & 42.7 \\
\bottomrule
\end{scoretabular}
\end{subtable}%
\begin{subtable}[t]{0.5\textwidth}
\centering
\footnotesize
\setlength{\tabcolsep}{2pt}
\caption{Bench2Drive-Geo (Town11+13)}
\label{tab:b2d-geo}
\begin{scoretabular}{@{}lccSE@{}}
\toprule
\multicolumn{1}{@{}c}{Training locations} & Eff. & Comfort & SR~$\uparrow$ & DS~$\uparrow$ \\
\midrule
Original (all cities) & 155.8 & 36.6 & 24.1 & 61.0 \\
w/o Town11+13 & 165.6 & 30.2 & 27.7 & 61.2 \\
\bottomrule
\end{scoretabular}
\end{subtable}
\end{table}


\section{Discussion}
\label{sec:discussion}
\subsection{Interpretation}
In both NAVSIM benchmarks we see that the recorded ego status reveals whether critical actions like braking are already initiated. Memory supplies the necessary road context, and the planner can then choose a conservative continuation. Although ego status is an intended input, over a short test horizon it can partly reveal the logged maneuver. High scores in collision- and traffic-light-related metrics therefore do not necessarily show that a model identified the current actors or signal states. And since our setting already reaches 91.1 PDMS on NAVSIMv1, the benchmark has little remaining score headroom to reward additional interaction capabilities.

In Bench2Drive, with the same underlying planner, removing access to dynamic information substantially lowers the final benchmark performance. This is the diagnostic gap our audit seeks to expose, and Bench2Drive moves toward it by evaluating closed-loop behavior over routes much longer than NAVSIM, a kind of test the research community should emphasize more. Its longer routes create more opportunities for collisions with other agents and reduce the dominant influence of the ego status. While NAVSIMv2 extends the evaluation horizon with simulated futures for an additional $\SI{4}{\second}$, those futures still start from rule-compliant waypoints with legitimate ego status. This reduces the chance that model errors compound into the kinds of closed-loop failures measured in Bench2Drive.

Since our HUGSIM-nuScenes results show that a long test horizon alone is insufficient, future benchmarks should also include counterfactual actors, as in RealEngine, to require necessary interactions and to better assess deployment-relevant driving capabilities.

\subsection{Limitations}
\label{sec:limitations}

Our protocol assumes repeated traversals of the same locations and high relative pose accuracy across traversals. This precludes evaluation on benchmarks such as WOD-E2E~\citep{xu2026wode2e}, where poses are not provided. When absolute ego poses are available, however, a related audit could retrieve street-level views and encode them with the same perception backbone into latent memories, following the spatial retrieval paradigm of \citet{jia2026spatial}. Such a variant would extend the audit to datasets without repeated logged traversals, while shifting the main caveats to pose alignment and domain gap.
To test whether our approach can be applied to benchmarks with less precise localization or fewer traversals, App.~\ref{app:sensitivity-tests} provides a sensitivity test. The results suggest that MemoryDrivoR remains useful under consumer-grade pose accuracy and with fewer traversals, albeit with a weaker diagnostic signal.


\section{Conclusion}

We introduced MemoryDrivoR as a new approach for auditing end-to-end driving benchmarks by substituting camera input from the evaluated scene with memory from previous drives. On NAVSIM, this setup achieves surprisingly high performance, competitive with leading camera-based methods despite relying only on ego status and static environmental information.
The Bench2Drive result qualifies this finding. There, evaluation runs over longer closed-loop routes with scenarios designed around traffic interactions, and the same intervention becomes much less successful.

Importantly, MemoryDrivoR does not establish the limit of what ego status and static information can support. Future work with better retrieval, richer memory representations, and more capable planner architectures could make it a sharper tool for benchmark auditing.

\newpage

\bibliographystyle{plainnat}
\begingroup
\hypersetup{urlcolor=black}
\bibliography{references}

@inproceedings{kirby2026drivor,
  author = {Kirby, Ellington and Boulch, Alexandre and Xu, Yihong and Yin, Yuan and Puy, Gilles and Zablocki, {\'E}loi and Bursuc, Andrei and Gidaris, Spyros and Marlet, Renaud and Bartoccioni, Florent and Cao, Anh-Quan and Samet, Nermin and Vu, Tuan-Hung and Cord, Matthieu},
  title = {Driving on Registers},
  booktitle = {Proceedings of the IEEE/CVF Conference on Computer Vision and Pattern Recognition (CVPR)},
  year = {2026}
}

@inproceedings{darcet2024vision,
  author = {Darcet, Timoth{\'e}e and Oquab, Maxime and Mairal, Julien and Bojanowski, Piotr},
  title = {Vision Transformers Need Registers},
  booktitle = {International Conference on Learning Representations (ICLR)},
  year = {2024}
}

@inproceedings{liao2025diffusiondrive,
  author = {Liao, Bencheng and Chen, Shaoyu and Yin, Haoran and Jiang, Bo and Wang, Cheng and Yan, Sixu and Zhang, Xinbang and Li, Xiangyu and Zhang, Ying and Zhang, Qian and Wang, Xinggang},
  title = {{DiffusionDrive}: Truncated Diffusion Model for End-to-End Autonomous Driving},
  booktitle = {Proceedings of the IEEE/CVF Conference on Computer Vision and Pattern Recognition (CVPR)},
  year = {2025}
}

@inproceedings{zhou2025autovla,
  author = {Zhou, Zewei and Cai, Tianhui and Zhao, Seth Z. and Zhang, Yun and Huang, Zhiyu and Zhou, Bolei and Ma, Jiaqi},
  title = {{AutoVLA}: A Vision-Language-Action Model for End-to-End Autonomous Driving with Adaptive Reasoning and Reinforcement Fine-Tuning},
  booktitle = {Advances in Neural Information Processing Systems (NeurIPS)},
  year = {2025}
}

@inproceedings{li2026recogdrive,
  author = {Li, Yongkang and Xiong, Kaixin and Guo, Xiangyu and Li, Fang and Yan, Sixu and Xu, Gangwei and Zhou, Lijun and Chen, Long and Sun, Haiyang and Wang, Bing and Ma, Kun and Chen, Guang and Ye, Hangjun and Liu, Wenyu and Wang, Xinggang},
  title = {{ReCogDrive}: A Reinforced Cognitive Framework for End-to-End Autonomous Driving},
  booktitle = {International Conference on Learning Representations (ICLR)},
  year = {2026}
}

@article{hu2025vlaad,
  author = {Hu, Tianshuai and Liu, Xiaolu and Wang, Song and Zhu, Yiyao and Liang, Ao and Kong, Lingdong and Zhao, Guoyang and Gong, Zeying and Cen, Jun and Huang, Zhiyu and Hao, Xiaoshuai and Li, Linfeng and Song, Hang and Li, Xiangtai and Ma, Jun and Shen, Shaojie and Zhu, Jianke and Tao, Dacheng and Liu, Ziwei and Liang, Junwei},
  title = {Vision-Language-Action Models for Autonomous Driving: Past, Present, and Future},
  journal = {arXiv preprint arXiv:2512.16760},
  year = {2025}
}

@inproceedings{feng2026rap,
  author = {Feng, Lan and Gao, Yang and Zablocki, {\'E}loi and Li, Quanyi and Li, Wuyang and Liu, Sichao and Cord, Matthieu and Alahi, Alexandre},
  title = {{RAP}: {3D} Rasterization Augmented End-to-End Planning},
  booktitle = {International Conference on Learning Representations (ICLR)},
  year = {2026}
}

@inproceedings{yao2026drivesuprim,
  author = {Yao, Wenhao and Li, Zhenxin and Lan, Shiyi and Wang, Zi and Sun, Xinglong and Alvarez, Jose M. and Wu, Zuxuan},
  title = {{DriveSuprim}: Towards Precise Trajectory Selection for End-to-End Planning},
  booktitle = {Proceedings of the AAAI Conference on Artificial Intelligence},
  year = {2026}
}

@inproceedings{nguyen2026lead,
  author = {Nguyen, Long and Fauth, Micha and Jaeger, Bernhard and Dauner, Daniel and Igl, Maximilian and Geiger, Andreas and Chitta, Kashyap},
  title = {{LEAD}: Minimizing Learner-Expert Asymmetry in End-to-End Driving},
  booktitle = {Proceedings of the IEEE/CVF Conference on Computer Vision and Pattern Recognition (CVPR)},
  year = {2026}
}

@article{li2025ztrs,
  author = {Li, Zhenxin and Yao, Wenhao and Wang, Zi and Sun, Xinglong and Chen, Jingde and Chang, Nadine and Shen, Maying and Song, Jingyu and Wu, Zuxuan and Lan, Shiyi and Alvarez, Jose M.},
  title = {{ZTRS}: Zero-Imitation End-to-end Autonomous Driving with Trajectory Scoring},
  journal = {arXiv preprint arXiv:2510.24108},
  year = {2025}
}

@inproceedings{renz2025simlingo,
  author = {Renz, Katrin and Chen, Long and Arani, Elahe and Sinavski, Oleg},
  title = {{SimLingo}: Vision-Only Closed-Loop Autonomous Driving with Language-Action Alignment},
  booktitle = {Proceedings of the IEEE/CVF Conference on Computer Vision and Pattern Recognition (CVPR)},
  year = {2025}
}

@inproceedings{hu2023planning,
  author = {Hu, Yihan and Yang, Jiazhi and Chen, Li and Li, Keyu and Sima, Chonghao and Zhu, Xizhou and Chai, Siqi and Du, Senyao and Lin, Tianwei and Wang, Wenhai and Lu, Lewei and Jia, Xiaosong and Liu, Qiang and Dai, Jifeng and Qiao, Yu and Li, Hongyang},
  title = {Planning-Oriented Autonomous Driving},
  booktitle = {Proceedings of the IEEE/CVF Conference on Computer Vision and Pattern Recognition (CVPR)},
  year = {2023}
}

@inproceedings{li2024think2drive,
  author = {Li, Qifeng and Jia, Xiaosong and Wang, Shaobo and Yan, Junchi},
  title = {{Think2Drive}: Efficient Reinforcement Learning by Thinking in Latent World Model for Quasi-Realistic Autonomous Driving (in {CARLA}-v2)},
  booktitle = {European Conference on Computer Vision (ECCV)},
  year = {2024}
}

@inproceedings{yang2025raw2drive,
  author = {Yang, Zhenjie and Jia, Xiaosong and Li, Qifeng and Yang, Xue and Yao, Maoqing and Yan, Junchi},
  title = {{Raw2Drive}: Reinforcement Learning with Aligned World Models for End-to-End Autonomous Driving (in {CARLA} v2)},
  booktitle = {Advances in Neural Information Processing Systems (NeurIPS)},
  year = {2025}
}

@article{zhai2023rethinking,
  author = {Zhai, Jiang-Tian and Feng, Ze and Du, Jinhao and Mao, Yongqiang and Liu, Jiang-Jiang and Tan, Zichang and Zhang, Yifu and Ye, Xiaoqing and Wang, Jingdong},
  title = {Rethinking the Open-Loop Evaluation of End-to-End Autonomous Driving in {nuScenes}},
  journal = {arXiv preprint arXiv:2305.10430},
  year = {2023}
}

@inproceedings{caesar2020nuscenes,
  author = {Caesar, Holger and Bankiti, Varun and Lang, Alex H. and Vora, Sourabh and Liong, Venice Erin and Xu, Qiang and Krishnan, Anush and Pan, Yu and Baldan, Giancarlo and Beijbom, Oscar},
  title = {{nuScenes}: A Multimodal Dataset for Autonomous Driving},
  booktitle = {Proceedings of the IEEE/CVF Conference on Computer Vision and Pattern Recognition (CVPR)},
  year = {2020}
}

@inproceedings{li2024egostatus,
  author = {Li, Zhiqi and Yu, Zhiding and Lan, Shiyi and Li, Jiahan and Kautz, Jan and Lu, Tong and Alvarez, Jose M.},
  title = {Is Ego Status All You Need for Open-Loop End-to-End Autonomous Driving?},
  booktitle = {Proceedings of the IEEE/CVF Conference on Computer Vision and Pattern Recognition (CVPR)},
  year = {2024}
}

@inproceedings{jia2024bench2drive,
  author = {Jia, Xiaosong and Yang, Zhenjie and Li, Qifeng and Zhang, Zhiyuan and Yan, Junchi},
  title = {{Bench2Drive}: Towards Multi-Ability Benchmarking of Closed-Loop End-To-End Autonomous Driving},
  booktitle = {Advances in Neural Information Processing Systems (NeurIPS), Datasets and Benchmarks Track},
  year = {2024}
}

@inproceedings{prakash2021multimodal,
  author = {Prakash, Aditya and Chitta, Kashyap and Geiger, Andreas},
  title = {Multi-Modal Fusion Transformer for End-to-End Autonomous Driving},
  booktitle = {Proceedings of the IEEE/CVF Conference on Computer Vision and Pattern Recognition (CVPR)},
  year = {2021}
}

@article{chitta2023transfuser,
  author = {Chitta, Kashyap and Prakash, Aditya and Jaeger, Bernhard and Yu, Zehao and Renz, Katrin and Geiger, Andreas},
  title = {{TransFuser}: Imitation With Transformer-Based Sensor Fusion for Autonomous Driving},
  journal = {IEEE Transactions on Pattern Analysis and Machine Intelligence},
  volume = {45},
  number = {11},
  pages = {12878--12895},
  year = {2023}
}

@inproceedings{dosovitskiy2017carla,
  author = {Dosovitskiy, Alexey and Ros, German and Codevilla, Felipe and Lopez, Antonio and Koltun, Vladlen},
  title = {{CARLA}: An Open Urban Driving Simulator},
  booktitle = {Proceedings of the Conference on Robot Learning (CoRL)},
  year = {2017}
}

@misc{carlaLeaderboard20Evaluation,
  author = {{CARLA Autonomous Driving Leaderboard}},
  title = {{CARLA} Leaderboard 2.0: Evaluation Criteria},
  howpublished = {\url{https://leaderboard.carla.org/evaluation_v2_0/}},
  year = {2025},
  note = {Accessed May 6, 2026}
}

@inproceedings{xu2026wode2e,
  author = {Xu, Runsheng and Lin, Hubert and Jeon, Wonseok and Feng, Hao and Zou, Yuliang and Sun, Liting and Gorman, John and Tolstaya, Kate and Tang, Sarah and White, Brandyn and Sapp, Ben and Tan, Mingxing and Hwang, Jyh-Jing and Anguelov, Dragomir},
  title = {{WOD-E2E}: {Waymo Open Dataset} for End-to-End Driving in Challenging Long-tail Scenarios},
  booktitle = {Proceedings of the IEEE/CVF Conference on Computer Vision and Pattern Recognition (CVPR)},
  year = {2026}
}

@inproceedings{sun2020waymo,
  author = {Sun, Pei and Kretzschmar, Henrik and Dotiwalla, Xerxes and Chouard, Aurelien and Patnaik, Vijaysai and Tsui, Paul and Guo, James and Zhou, Yin and Chai, Yuning and Caine, Benjamin and Vasudevan, Vijay and Han, Wei and Ngiam, Jiquan and Zhao, Hang and Timofeev, Aleksei and Ettinger, Scott and Krivokon, Maxim and Gao, Amy and Joshi, Aditya and Zhao, Sheng and Cheng, Shuyang and Zhang, Yu and Shlens, Jonathon and Chen, Zhifeng and Anguelov, Dragomir},
  title = {Scalability in Perception for Autonomous Driving: {Waymo Open Dataset}},
  booktitle = {Proceedings of the IEEE/CVF Conference on Computer Vision and Pattern Recognition (CVPR)},
  year = {2020}
}

@misc{openscene2023,
  author = {{OpenScene Contributors}},
  title = {{OpenScene}: The Largest Up-to-Date {3D} Occupancy Prediction Benchmark in Autonomous Driving},
  howpublished = {\url{https://github.com/OpenDriveLab/OpenScene}},
  year = {2023}
}

@inproceedings{dauner2024navsim,
  author = {Dauner, Daniel and Hallgarten, Marcel and Li, Tianyu and Weng, Xinshuo and Huang, Zhiyu and Yang, Zetong and Li, Hongyang and Gilitschenski, Igor and Ivanovic, Boris and Pavone, Marco and Geiger, Andreas and Chitta, Kashyap},
  title = {{NAVSIM}: Data-Driven Non-Reactive Autonomous Vehicle Simulation and Benchmarking},
  booktitle = {Advances in Neural Information Processing Systems (NeurIPS), Datasets and Benchmarks Track},
  year = {2024}
}

@inproceedings{dauner2023parting,
  author = {Dauner, Daniel and Hallgarten, Marcel and Geiger, Andreas and Chitta, Kashyap},
  title = {Parting with Misconceptions about Learning-based Vehicle Motion Planning},
  booktitle = {Proceedings of the Conference on Robot Learning (CoRL)},
  year = {2023}
}

@inproceedings{karnchanachari2024nuplan,
  author = {Karnchanachari, Napat and Geromichalos, Dimitris and Tan, Kok Seang and Li, Nanxiang and Eriksen, Christopher and Yaghoubi, Shakiba and Mehdipour, Noushin and Bernasconi, Gianmarco and Fong, Whye Kit and Guo, Yiluan and Caesar, Holger},
  title = {Towards Learning-Based Planning: The {nuPlan} Benchmark for Real-World Autonomous Driving},
  booktitle = {Proceedings of the IEEE International Conference on Robotics and Automation (ICRA)},
  year = {2024}
}

@inproceedings{zeng2026priordrive,
  author = {Zeng, Shuang and Chang, Xinyuan and Liu, Xinran and Yuan, Yujian and Liang, Shiyi and Pan, Zheng and Xu, Mu and Wei, Xing},
  title = {{PriorDrive}: Enhancing Online {HD} Mapping with Unified Vector Priors},
  booktitle = {Proceedings of the AAAI Conference on Artificial Intelligence},
  year = {2026}
}

@inproceedings{devlin2019bert,
  author = {Devlin, Jacob and Chang, Ming-Wei and Lee, Kenton and Toutanova, Kristina},
  title = {{BERT}: Pre-training of Deep Bidirectional Transformers for Language Understanding},
  booktitle = {Proceedings of the North American Chapter of the Association for Computational Linguistics: Human Language Technologies (NAACL-HLT)},
  year = {2019}
}

@inproceedings{xiong2023neural,
  author = {Xiong, Xuan and Liu, Yicheng and Yuan, Tianyuan and Wang, Yue and Wang, Yilun and Zhao, Hang},
  title = {Neural Map Prior for Autonomous Driving},
  booktitle = {Proceedings of the IEEE/CVF Conference on Computer Vision and Pattern Recognition (CVPR)},
  year = {2023}
}

@inproceedings{yuan2024streammapnet,
  author = {Yuan, Tianyuan and Liu, Yicheng and Wang, Yue and Wang, Yilun and Zhao, Hang},
  title = {{StreamMapNet}: Streaming Mapping Network for Vectorized Online {HD} Map Construction},
  booktitle = {Proceedings of the IEEE/CVF Winter Conference on Applications of Computer Vision (WACV)},
  year = {2024}
}

@inproceedings{lilja2024localization,
  author = {Lilja, Adam and Fu, Junsheng and Stenborg, Erik and Hammarstrand, Lars},
  title = {Localization Is All You Evaluate: Data Leakage in Online Mapping Datasets and How to Fix It},
  booktitle = {Proceedings of the IEEE/CVF Conference on Computer Vision and Pattern Recognition (CVPR)},
  year = {2024}
}

@inproceedings{you2022hindsight,
  author = {You, Yurong and Luo, Katie Z. and Chen, Xiangyu and Chen, Junan and Chao, Wei-Lun and Sun, Wen and Hariharan, Bharath and Campbell, Mark and Weinberger, Kilian Q.},
  title = {Hindsight Is 20/20: Leveraging Past Traversals to Aid {3D} Perception},
  booktitle = {International Conference on Learning Representations (ICLR)},
  year = {2022}
}

@inproceedings{yuan2024presight,
  author = {Yuan, Tianyuan and Mao, Yucheng and Yang, Jiawei and Liu, Yicheng and Wang, Yue and Zhao, Hang},
  title = {{PreSight}: Enhancing Autonomous Vehicle Perception with City-Scale {NeRF} Priors},
  booktitle = {European Conference on Computer Vision (ECCV)},
  year = {2024}
}

@inproceedings{li2024memorize,
  author = {Li, Yiming and Wang, Zehong and Wang, Yue and Yu, Zhiding and Gojcic, Zan and Pavone, Marco and Feng, Chen and Alvarez, Jose M.},
  title = {Memorize What Matters: Emergent Scene Decomposition from Multitraverse},
  booktitle = {Advances in Neural Information Processing Systems (NeurIPS)},
  year = {2024}
}

@article{zhou2026compressed,
  author = {Zhou, Brady and Kr{\"a}henb{\"u}hl, Philipp},
  title = {Compressed Map Priors for {3D} Perception},
  journal = {arXiv preprint arXiv:2601.00139},
  year = {2026}
}

@article{li2024hydra,
  author = {Li, Zhenxin and Li, Kailin and Wang, Shihao and Lan, Shiyi and Yu, Zhiding and Ji, Yishen and Li, Zhiqi and Zhu, Ziyue and Kautz, Jan and Wu, Zuxuan and Jiang, Yu-Gang and Alvarez, Jose M.},
  title = {{Hydra-MDP}: End-to-End Multimodal Planning with Multi-Target {Hydra-Distillation}},
  journal = {arXiv preprint arXiv:2406.06978},
  year = {2024}
}

@inproceedings{cao2025pseudosimulation,
  author = {Cao, Wei and Hallgarten, Marcel and Li, Tianyu and Dauner, Daniel and Gu, Xunjiang and Wang, Caojun and Miron, Yakov and Aiello, Marco and Li, Hongyang and Gilitschenski, Igor and Ivanovic, Boris and Pavone, Marco and Geiger, Andreas and Chitta, Kashyap},
  title = {Pseudo-Simulation for Autonomous Driving},
  booktitle = {Proceedings of the Conference on Robot Learning (CoRL)},
  year = {2025}
}

@inproceedings{jia2026spatial,
  author = {Jia, Xiaosong and Zhang, Chenhe and Jiang, Yule and Wong, Songbur and Zhang, Zhiyuan and Chen, Chen and Zhang, Shaofeng and Zhou, Xuanhe and Yang, Xue and Yan, Junchi and Jiang, Yu-Gang},
  title = {Spatial Retrieval Augmented Autonomous Driving},
  booktitle = {Proceedings of the IEEE/CVF Conference on Computer Vision and Pattern Recognition (CVPR)},
  year = {2026}
}

@inproceedings{yeon2026prioreye,
  author = {Yeon, Kyuhwan and Ramtoula, Benjamin and De Martini, Daniele},
  title = {{PriorEye}: Geospatial Visual Priors for End-to-End Autonomous Driving},
  booktitle = {European Conference on Computer Vision (ECCV)},
  year = {2026}
}

@inproceedings{alayrac2022flamingo,
  author = {Alayrac, Jean-Baptiste and Donahue, Jeff and Luc, Pauline and Miech, Antoine and Barr, Iain and Hasson, Yana and Lenc, Karel and Mensch, Arthur and Millican, Katherine and Reynolds, Malcolm and Ring, Roman and Rutherford, Eliza and Cabi, Serkan and Han, Tengda and Gong, Zhitao and Samangooei, Sina and Monteiro, Marianne and Menick, Jacob and Borgeaud, Sebastian and Brock, Andrew and Nematzadeh, Aida and Sharifzadeh, Sahand and Binkowski, Mikolaj and Barreira, Ricardo and Vinyals, Oriol and Zisserman, Andrew and Simonyan, Karen},
  title = {{Flamingo}: A Visual Language Model for Few-Shot Learning},
  booktitle = {Advances in Neural Information Processing Systems (NeurIPS)},
  year = {2022}
}

@inproceedings{mildenhall2020nerf,
  author = {Mildenhall, Ben and Srinivasan, Pratul P. and Tancik, Matthew and Barron, Jonathan T. and Ramamoorthi, Ravi and Ng, Ren},
  title = {{NeRF}: Representing Scenes as Neural Radiance Fields for View Synthesis},
  booktitle = {European Conference on Computer Vision (ECCV)},
  year = {2020}
}

@inproceedings{tancik2020fourier,
  author = {Tancik, Matthew and Srinivasan, Pratul P. and Mildenhall, Ben and Fridovich-Keil, Sara and Raghavan, Nithin and Singhal, Utkarsh and Ramamoorthi, Ravi and Barron, Jonathan T. and Ng, Ren},
  title = {Fourier Features Let Networks Learn High Frequency Functions in Low Dimensional Domains},
  booktitle = {Advances in Neural Information Processing Systems (NeurIPS)},
  year = {2020}
}

@misc{agc2025navhardleaderboard,
  author = {{Autonomous Grand Challenge}},
  title = {{NAVSIM} v2 {NavHard} End-to-End Driving Leaderboard},
  howpublished = {\url{https://huggingface.co/spaces/AGC2025/e2e-driving-navhard}},
  year = {2025},
  note = {Public leaderboard on Hugging Face, accessed April 22, 2026}
}

@misc{agc2024navtestleaderboard,
  author = {{Autonomous Grand Challenge}},
  title = {{NAVSIM} v1 {Navtest} End-to-End Driving Leaderboard},
  howpublished = {\url{https://huggingface.co/spaces/AGC2024-P/e2e-driving-navtest}},
  year = {2024},
  note = {Public leaderboard on Hugging Face, accessed April 22, 2026}
}

@misc{drivorRepository,
  author = {{DrivoR Contributors}},
  title = {{DrivoR} Repository, Model Weights, and License Statement},
  howpublished = {\url{https://github.com/valeoai/DrivoR}},
  year = {2026},
  note = {Accessed May 6, 2026}
}

@misc{navsimRepository,
  author = {{NAVSIM Contributors}},
  title = {{NAVSIM} Repository and License Statement},
  howpublished = {\url{https://github.com/autonomousvision/navsim}},
  year = {2026},
  note = {Accessed May 6, 2026}
}

@misc{opensceneRepository,
  author = {{OpenScene Contributors}},
  title = {{OpenScene} Repository and License Statement},
  howpublished = {\url{https://github.com/OpenDriveLab/OpenScene}},
  year = {2026},
  note = {Accessed May 6, 2026}
}

@misc{nuplanDevkit,
  author = {{Motional}},
  title = {{nuPlan} Devkit Repository and License Statement},
  howpublished = {\url{https://github.com/motional/nuplan-devkit}},
  year = {2026},
  note = {Accessed May 6, 2026}
}

@misc{bench2driveRepository,
  author = {{Bench2Drive Contributors}},
  title = {{Bench2Drive} Repository and License Statement},
  howpublished = {\url{https://github.com/Thinklab-SJTU/Bench2Drive}},
  year = {2026},
  note = {Accessed May 6, 2026}
}

@misc{bench2driveHFDataset,
  author = {{SJTU ReThinkLab}},
  title = {{Bench2Drive} Dataset Card},
  howpublished = {\url{https://huggingface.co/datasets/rethinklab/Bench2Drive}},
  year = {2026},
  note = {Accessed May 6, 2026}
}

@misc{bench2driveFullHFDataset,
  author = {{SJTU ReThinkLab}},
  title = {{Bench2Drive-Full} Dataset Card},
  howpublished = {\url{https://huggingface.co/datasets/rethinklab/Bench2Drive-Full}},
  year = {2026},
  note = {Accessed May 6, 2026}
}

@misc{carla0915Repository,
  author = {{CARLA Simulator Contributors}},
  title = {{CARLA} 0.9.15 Repository License Statement},
  howpublished = {\url{https://github.com/carla-simulator/carla/tree/0.9.15}},
  year = {2026},
  note = {Accessed May 6, 2026}
}

@misc{carla0915Release,
  author = {{CARLA Simulator Contributors}},
  title = {{CARLA} 0.9.15 Release},
  howpublished = {\url{https://github.com/carla-simulator/carla/releases/tag/0.9.15}},
  year = {2023},
  note = {Accessed May 6, 2026}
}

@misc{realengineRepository,
  author = {{RealEngine Contributors}},
  title = {{RealEngine} Repository and License Statement},
  howpublished = {\url{https://github.com/fudan-zvg/RealEngine}},
  year = {2026},
  note = {Accessed August 28, 2026}
}

@misc{streetGaussiansLicense,
  author = {{Street Gaussians Contributors}},
  title = {{Street Gaussians} License Statement},
  howpublished = {\url{https://github.com/zju3dv/street_gaussians/blob/main/LICENSE}},
  year = {2026},
  note = {Accessed August 28, 2026}
}

@misc{hugsimRepository,
  author = {{HUGSIM Contributors}},
  title = {{HUGSIM} Repository and License Statement},
  howpublished = {\url{https://github.com/hyzhou404/HUGSIM}},
  year = {2026},
  note = {Accessed August 28, 2026}
}

@misc{hugsimHFDataset,
  author = {{X-Dimensional Representations Lab}},
  title = {{HUGSIM} Dataset Card},
  howpublished = {\url{https://huggingface.co/datasets/XDimLab/HUGSIM}},
  year = {2026},
  note = {Accessed August 28, 2026}
}

@misc{threeDRealCarRepository,
  author = {{3DRealCar Contributors}},
  title = {{3DRealCar} Toolkit Repository and Dataset License Statement},
  howpublished = {\url{https://github.com/xiaobiaodu/3DRealCar_Toolkit}},
  year = {2026},
  note = {Accessed August 28, 2026}
}

@misc{nuscenesTerms,
  author = {{Motional}},
  title = {{nuScenes} Dataset Terms},
  howpublished = {\url{https://www.nuscenes.org/terms-of-use}},
  year = {2021},
  note = {Accessed August 28, 2026}
}

@article{kerbl20233dgs,
  author = {Kerbl, Bernhard and Kopanas, Georgios and Leimk{\"u}hler, Thomas and Drettakis, George},
  title = {{3D} {Gaussian} Splatting for Real-Time Radiance Field Rendering},
  journal = {ACM Transactions on Graphics},
  volume = {42},
  number = {4},
  pages = {139:1--139:14},
  year = {2023}
}

@article{treiber2000idm,
  author = {Treiber, Martin and Hennecke, Ansgar and Helbing, Dirk},
  title = {Congested Traffic States in Empirical Observations and Microscopic Simulations},
  journal = {Physical Review E},
  volume = {62},
  number = {2},
  pages = {1805--1824},
  year = {2000}
}

@inproceedings{du2025rtmap,
  author = {Du, Yuheng and Yang, Sheng and Wang, Lingxuan and Hou, Zhenghua and Cai, Chengying and Tan, Zhitao and Chen, Mingxia and Huang, Shi-Sheng and Li, Qiang},
  title = {{RTMap}: Real-Time Recursive Mapping with Change Detection and Localization},
  booktitle = {Proceedings of the IEEE/CVF International Conference on Computer Vision (ICCV)},
  year = {2025}
}

@article{zhou2026hugsim,
  author = {Zhou, Hongyu and Lin, Longzhong and Wang, Jiabao and Lu, Yichong and Bai, Dongfeng and Liu, Bingbing and Wang, Yue and Geiger, Andreas and Liao, Yiyi},
  title = {{HUGSIM}: A Real-Time, Photo-Realistic and Closed-Loop Simulator for Autonomous Driving},
  journal = {IEEE Transactions on Pattern Analysis and Machine Intelligence},
  volume = {48},
  number = {4},
  pages = {4673--4691},
  year = {2026}
}

@article{jiang2025realengine,
  author = {Jiang, Junzhe and Song, Nan and Li, Jingyu and Zhu, Xiatian and Zhang, Li},
  title = {{RealEngine}: Simulating Autonomous Driving in Realistic Context},
  journal = {arXiv preprint arXiv:2505.16902},
  year = {2025}
}
\endgroup

\newpage
\appendix

\suppressfloats[t]
\section*{Appendix Overview}
This appendix collects supporting material for the discussion in the main paper. It reports additional results, including an analysis of specific failure scenarios, component ablations, and extra qualitative examples. It also provides implementation details for memory retrieval, the planner for Bench2Drive, and the HD-map baseline. It concludes with a discussion of societal impacts.

\section{Additional Results}

\subsection{Failure Analysis}
\label{app:failure-analysis}

Bench2Drive groups its 220 routes into five ability families: \emph{Merging}, \emph{Overtaking}, \emph{Emergency Brake}, \emph{Give Way}, and \emph{Traffic Sign}. Tab.~\ref{tab:appendix-b2d-ability-detailed} reports the corresponding scores for the methods considered above whenever per-family results are available. Removing camera input lowers performance across every ability family. Thus, even after separating routes by ability type, we do not find a category that can be solved from memory alone. MemoryDrivoR retains its strongest residual performance on \emph{Merging}, consistent with the HighwayCutIn result in Fig.~\ref{fig:b2d-same-route}, where it completes all 5 routes.

\begin{table}[t]
\centering
\setlength{\tabcolsep}{2pt}
\begin{threeparttable}
\small
\caption{\textbf{Bench2Drive} multi-ability breakdown.}
\label{tab:appendix-b2d-ability-detailed}
\begin{scoretabular}{lccN{2.1}N{2.1}N{2.1}N{2.1}N{2.1}M{1.0cm}@{}}
\toprule
\multicolumn{1}{c}{Method} & Cam. & Mem. & \multicolumn{1}{c}{Merge} & \multicolumn{1}{c}{Overtake} & \multicolumn{1}{c}{Emerg. Brake} & \multicolumn{1}{c}{Give Way} & \multicolumn{1}{c}{Traf. Sign} & \scorecell{Mean~$\uparrow$} \\
\midrule
\refrowcell{Think2Drive$^\ddagger$~\citep{li2024think2drive}} & \multicolumn{2}{c}{\refrowcell{privileged}} & \refnumcell{2.1}{81.3} & \refnumcell{2.1}{83.9} & \refnumcell{2.1}{90.2} & \refnumcell{2.1}{90.0} & \refnumcell{2.1}{87.7} & \refrowcell{\tablenum[table-format=2.1]{86.3}} \\
\midrule
UniAD$^\dagger$~\citep{hu2023planning} & \cmark & \xmark & 14.1 & 17.8 & 21.7 & 10.0 & 14.2 & \tablenum[table-format=2.1]{15.6} \\
ReCogDrive~\citep{li2026recogdrive} & \cmark & \xmark & 29.7 & 20.0 & 69.1 & 20.0 & 71.3 & \tablenum[table-format=2.1]{42.0} \\
SimLingo~\citep{renz2025simlingo} & \cmark & \xmark & \bfseries 60.0 & \bfseries 60.0 & \bfseries 78.3 & \bfseries 50.0 & \bfseries 77.9 & \bfseries\tablenum[table-format=2.1]{65.3} \\
DrivoR$^\ast$~\citep{kirby2026drivor} & \cmark & \xmark & 20.0 & 17.7 & 50.0 & \bfseries 50.0 & 40.5 & \tablenum[table-format=2.1]{35.7} \\
\midrule
AD-MLP$^\dagger$~\citep{zhai2023rethinking} & \xmark & \xmark & 0.0 & 0.0 & 0.0 & \bfseries 0.0 & 4.4 & \tablenum[table-format=2.1]{0.9} \\
DrivoR$^\ast$ & \xmark & \xmark & 5.6 & 0.0 & 0.0 & \bfseries 0.0 & 9.6 & \tablenum[table-format=2.1]{3.0} \\
MemoryDrivoR (ours) & \xmark & \cmark & \bfseries 13.8 & \bfseries 6.7 & \bfseries 11.7 & \bfseries 0.0 & \bfseries 13.2 & \bfseries\tablenum[table-format=2.1]{9.0} \\
\bottomrule
\end{scoretabular}
\begin{tablenotes}[flushleft]
\footnotesize
\item[] $^\dagger$ Results from~\citep{jia2024bench2drive}. $^\ddagger$ Results are from~\citep{yang2025raw2drive}. $^\ast$ Results are based on our own implementation.
\end{tablenotes}
\end{threeparttable}
\end{table}

Since those ability families are quite broad, we also grouped all 220 routes by disjoint interaction types. Tab.~\ref{tab:appendix-b2d-interactions} reports success rates and driving scores for MemoryDrivoR and camera-based DrivoR. While MemoryDrivoR has a lower score for every interaction type, we see that its losses are concentrated in interactions that require observing the current scenario. Among the types with at least 25 routes, the largest gaps occur for pedestrians and cyclists, cut-ins and sudden obstacles, and traffic-controlled junctions. Specifically, MemoryDrivoR produces more pedestrian collisions, collisions in cut-in/hard-braking scenarios, and red-light violations.

\begin{table}[t]
\centering
\setlength{\tabcolsep}{3pt}
\caption{\textbf{Bench2Drive failure analysis.} We report success rate and driving score.}
\label{tab:appendix-b2d-interactions}
\begin{scoretabular}{lcN{3.1}N{3.1}N{3.1}N{3.1}G{-2.1}}
\toprule
\multirow{2}{*}{Interaction} & \multirow{2}{*}{\#} & \multicolumn{2}{c}{MemoryDrivoR} & \multicolumn{2}{c}{DrivoR} & \multicolumn{1}{>{\columncolor{scoregray}}c}{\phantom{$\Delta$ DS}} \\
 & & \multicolumn{1}{c}{SR} & \multicolumn{1}{c}{DS} & \multicolumn{1}{c}{SR} & \multicolumn{1}{c}{DS} & \multicolumn{1}{>{\columncolor{scoregray}}c}{\multirow{-2}{*}{$\Delta$ DS}} \\
\midrule
Control loss due to poor road conditions & 5 & 60.0 & 60.0 & 100.0 & 100.0 & -40.0 \\
Emergency-vehicle yielding & 5 & 0.0 & 32.3 & 0.0 & 70.0 & -37.7 \\
Pedestrians and cyclists & 30 & 3.3 & 28.2 & 36.7 & 64.5 & -36.3 \\
Cut-ins, hard braking, and sudden obstacles & 25 & 28.0 & 47.8 & 68.0 & 82.0 & -34.2 \\
Signalized and traffic-controlled junctions & 45 & 4.4 & 29.5 & 26.7 & 62.7 & -33.2 \\
Obstacle avoidance and lane borrowing & 45 & 6.7 & 27.6 & 17.8 & 49.4 & -21.8 \\
Unsignalized junctions and turning conflicts & 30 & 3.3 & 34.9 & 23.3 & 54.4 & -19.5 \\
Merging, flow crossing, and lane changes & 35 & 11.4 & 43.2 & 20.0 & 54.6 & -11.4 \\
\bottomrule
\end{scoretabular}
\end{table}

For NAVSIM, we group all 12,146 NAVSIMv1 scenes and 450 NAVSIMv2 scenes by self-derived multi-label interaction types using the ground-truth traffic state. As shown in Tab.~\ref{tab:appendix-navsim-interactions}, the largest gaps occur for lead vehicles on NAVSIMv1 and vehicle merges on NAVSIMv2, while the smallest gaps occur for vehicle merges on NAVSIMv1 and red lights on NAVSIMv2. However, in contrast to Bench2Drive, no interaction type in NAVSIM produces a clear separation in favor of DrivoR. This suggests that these interaction types do not consistently demand responses to the current dynamic scenario that are strong enough to distinguish the two planners.

\begin{table}[t]
\centering
\setlength{\tabcolsep}{4pt}
\caption{\textbf{NAVSIM failure analysis} by interaction type.}
\label{tab:appendix-navsim-interactions}
\begin{scoretabular}{clcN{2.1}N{2.1}G{+1.1}}
\toprule
Benchmark & Interaction & \# & \multicolumn{1}{c}{MemoryDrivoR} & \multicolumn{1}{c}{DrivoR} & \multicolumn{1}{>{\columncolor{scoregray}}c}{$\Delta$} \\
\midrule
\multirow{5}{*}{\shortstack[c]{NAVSIMv1\\PDMS~$\uparrow$}} & Lead vehicle & 5,409 & 90.7 & 93.4 & -2.7 \\
 & Dense traffic & 5,915 & 91.2 & 93.6 & -2.4 \\
 & $\ldots$ & & & & \multicolumn{1}{>{\columncolor{scoregray}}c}{} \\
 & Bicyclist & 694 & 91.1 & 92.0 & -0.9 \\
 & Vehicle merge & 135 & 92.2 & 91.5 & +0.8 \\
\midrule
\multirow{5}{*}{\shortstack[c]{NAVSIMv2\\EPDMS~$\uparrow$}} & Vehicle merge & 44 & 46.8 & 53.0 & -6.2 \\
 & Pedestrian & 199 & 47.4 & 53.2 & -5.8 \\
 & $\ldots$ & & & & \multicolumn{1}{>{\columncolor{scoregray}}c}{} \\
 & Traffic-signal scene & 207 & 47.9 & 49.2 & -1.3 \\
 & Red light on route & 98 & 49.9 & 50.3 & -0.4 \\
\bottomrule
\end{scoretabular}
\end{table}

\subsection{Controlled Evaluation Horizon}
\label{app:controlled-horizon}
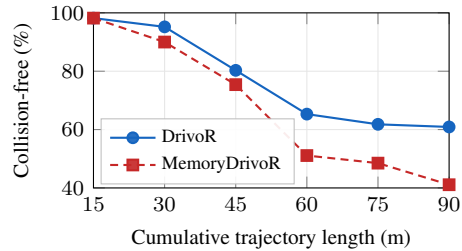
\begin{wrapfigure}{r}{0.45\textwidth}
\vspace{-2.2\baselineskip}
\centering
\begin{tikzpicture}
\begin{axis}[
    width=\linewidth,
    height=0.62\linewidth,
    xmin=15,
    xmax=90,
    ymin=40,
    ymax=100,
    xtick={15,30,45,60,75,90},
    ytick={40,60,80,100},
    xlabel={Cumulative trajectory length (m)},
    ylabel={Collision-free (\%)},
    grid=major,
    grid style={black!10},
    tick label style={font=\footnotesize},
    label style={font=\footnotesize},
    legend style={
        draw=black!25,
        fill=white,
        fill opacity=0.92,
        text opacity=1,
        at={(0.02,0.02)},
        anchor=south west,
        font=\scriptsize,
    },
    legend cell align=left,
]
\addplot+[plotblue, thick, mark=*, mark options={solid}] coordinates {
    (15,98.2) (30,95.2) (45,80.3) (60,65.3) (75,61.8) (90,60.9)
};
\addlegendentry{DrivoR}
\addplot+[plotred, thick, densely dashed, mark=square*, mark options={solid}] coordinates {
    (15,98.2) (30,90.0) (45,75.4) (60,51.1) (75,48.5) (90,41.1)
};
\addlegendentry{MemoryDrivoR}

\end{axis}
\end{tikzpicture}
\caption{\textbf{Ratio of collision-free routes} by traveled distance on Bench2Drive.}
\label{fig:appendix-controlled-horizon}
\vspace{-1\baselineskip}
\end{wrapfigure}
To quantify the influence of the evaluation horizon, we measure the ratio of routes where a model did not collide with any dynamic object as a function of the cumulative trajectory length per route. We use distance rather than time to avoid rewarding models that drive slowly. Fig.~\ref{fig:appendix-controlled-horizon} shows the plot for Bench2Drive. The gap widens overall with distance traveled, showing that the longer the route, the more likely MemoryDrivoR is to collide with other traffic participants relative to DrivoR.

\subsection{Detailed HUGSIM Results}
\label{app:additional-benchmark-details}
HUGSIM~\citep{zhou2026hugsim} is a photorealistic closed-loop simulator built from reconstructed real-world scenes with controllable traffic actors. Although it builds on four datasets, nuScenes is the only one with sufficient spatial overlap and pose information across traversals for our retrieval setup. HUGSIM requests a new plan every 0.25 seconds and terminates on collision, route deviation, completion, agent failure, or its step limit. We evaluate on all 87 nuScenes scenarios in HUGSIM based on the test protocol used by DrivoR. The full breakdown by difficulty can be found in Tab.~\ref{tab:appendix-hugsim-difficulty}.

\begin{table}[t]
\centering
\setlength{\tabcolsep}{3pt}
\begin{threeparttable}
\caption{\textbf{HUGSIM-nuScenes} HD-Score by difficulty. Brackets give scenario counts. Zero-shot evaluation of the DrivoR models trained on NAVSIMv1.}
\label{tab:appendix-hugsim-difficulty}
\begin{scoretabular}{lccccccS}
\toprule
Method & Cam. & Mem. & Easy [12] & Medium [33] & Hard [19] & Extreme [23] & \scorecell{Overall [87]} \\
\midrule
UniAD$^\dagger$ & \cmark & \xmark & 75.9 & \textbf{46.4} & 29.9 & 7.9 & \textbf{36.7} \\
DrivoR & \cmark & \xmark & \textbf{84.6} & 18.0 & \textbf{32.0} & \textbf{41.1} & 36.4 \\
\midrule
DrivoR & \xmark & \xmark & 59.2 & \textbf{19.7} & \textbf{22.1} & 32.4 & 29.1 \\
MemoryDrivoR & \xmark & \cmark & \textbf{83.6} & 19.0 & 17.1 & \textbf{36.9} & \textbf{32.2} \\
\bottomrule
\end{scoretabular}
\begin{tablenotes}[flushleft]
\footnotesize
\item[] $^\dagger$ Trained on nuScenes.
\end{tablenotes}
\end{threeparttable}
\end{table}

\subsection{Sensitivity Tests}
\label{app:sensitivity-tests}
\label{app:pose-noise}
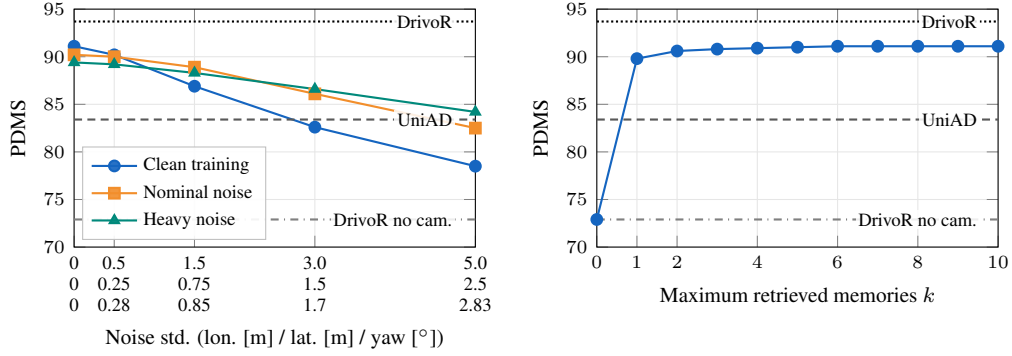
\begin{figure}[t]
  \centering
  \begin{tikzpicture}
\begin{groupplot}[
    group style={
        group size=2 by 1,
        horizontal sep=0.115\textwidth,
    },
    width=0.38\textwidth,
    height=0.225\textwidth,
    scale only axis,
    ymin=70,
    ymax=95,
    ytick={70,75,80,85,90,95},
    grid=major,
    grid style={black!10},
    tick label style={font=\scriptsize},
    label style={font=\footnotesize},
]
\nextgroupplot[
    xlabel={Noise std. (lon. [m] / lat. [m] / yaw [$^\circ$])},
    ylabel={PDMS},
    xmin=0,
    xmax=5,
    xtick={0,0.5,1.5,3,5},
    xticklabels={
        \shortstack{0\\0\\0},
        \shortstack{0.5\\0.25\\0.28},
        \shortstack{1.5\\0.75\\0.85},
        \shortstack{3.0\\1.5\\1.7},
        \shortstack{5.0\\2.5\\2.83}
    },
    xticklabel style={align=center, font=\scriptsize},
    legend style={draw=black!25, fill=white, fill opacity=0.92, text opacity=1, at={(0.02,0.03)}, anchor=south west, font=\scriptsize},
    legend cell align=left,
]
\addplot+[plotblue, thick, mark=*, mark options={solid}] coordinates {
    (0,91.1) (0.5,90.2) (1.5,86.9) (3,82.6) (5,78.5)
};
\addlegendentry{Clean training}
\addplot+[plotorange, thick, mark=square*, mark options={solid}] coordinates {
    (0,90.2) (0.5,90.0) (1.5,88.9) (3,86.1) (5,82.5)
};
\addlegendentry{Nominal noise}
\addplot+[plotgreen, thick, mark=triangle*, mark options={solid}] coordinates {
    (0,89.4) (0.5,89.2) (1.5,88.3) (3,86.6) (5,84.2)
};
\addlegendentry{Heavy noise}
\addplot[black!65, densely dashed, thick, forget plot] coordinates {
    (0,83.4) (5,83.4)
};
\addplot[black!50, dashdotted, thick, forget plot] coordinates {
    (0,72.9) (5,72.9)
};
\addplot[black, densely dotted, thick, forget plot] coordinates {
    (0,93.7) (5,93.7)
};
\node[anchor=east, font=\scriptsize, fill=white, fill opacity=0.86, text opacity=1, inner sep=1pt] at (axis cs:4.75,93.7) {DrivoR};
\node[anchor=east, font=\scriptsize, fill=white, fill opacity=0.86, text opacity=1, inner sep=1pt] at (axis cs:4.75,83.4) {UniAD};
\node[anchor=east, font=\scriptsize, fill=white, fill opacity=0.86, text opacity=1, inner sep=1pt] at (axis cs:4.75,72.9) {DrivoR no cam.};

\nextgroupplot[
    xlabel={Maximum retrieved memories $k$},
    ylabel={PDMS},
    xmin=0,
    xmax=10,
    xtick={0,1,2,4,6,8,10},
]
\addplot+[plotblue, thick, mark=*, mark options={solid}] coordinates {
    (0,72.9)
    (1,89.8)
    (2,90.6)
    (3,90.8)
    (4,90.9)
    (5,91.0)
    (6,91.1)
    (7,91.1)
    (8,91.1)
    (9,91.1)
    (10,91.1)
};
\addplot[black, densely dotted, thick, forget plot] coordinates {
    (0,93.7) (10,93.7)
};
\addplot[black!65, densely dashed, thick, forget plot] coordinates {
    (0,83.4) (10,83.4)
};
\addplot[black!50, dashdotted, thick, forget plot] coordinates {
    (0,72.9) (10,72.9)
};
\node[anchor=east, font=\scriptsize, fill=white, fill opacity=0.86, text opacity=1, inner sep=1pt] at (axis cs:9.55,93.7) {DrivoR};
\node[anchor=east, font=\scriptsize, fill=white, fill opacity=0.86, text opacity=1, inner sep=1pt] at (axis cs:9.55,83.4) {UniAD};
\node[anchor=east, font=\scriptsize, fill=white, fill opacity=0.86, text opacity=1, inner sep=1pt] at (axis cs:9.55,72.9) {DrivoR no cam.};
\end{groupplot}
\end{tikzpicture}
  \caption{\textbf{Sensitivity tests} for MemoryDrivoR on NAVSIMv1. Left: pose-noise robustness. The three x-axis rows report longitudinal, lateral, and yaw noise standard deviations during evaluation. Right: sensitivity to the maximum number of retrieved memories. The labeled horizontal reference lines show DrivoR, UniAD, and the no-camera DrivoR baseline.}
  \label{fig:appendix-sensitivity-navsim-v1}
  \label{fig:appendix-pose-noise-navsim-v1}
  \label{fig:appendix-k-sensitivity-navsim-v1}
\end{figure}

The main experiments use accurate global poses and \(k=10\) retrieved memories to isolate how much benchmark performance is achievable from static information alone. This setting gives the cleanest diagnostic, but the audit should also tolerate less accurate poses and fewer repeated traversals to be widely applicable. Fig.~\ref{fig:appendix-sensitivity-navsim-v1} probes this with two sensitivity tests. The first perturbs the poses used for memory construction and retrieval. The second restricts the number of retrieved memories.

\paragraph{Inaccurate poses.} For the pose-noise check, the perturbations are Gaussian and applied to both the current ego pose and the memory-bank poses at evaluation time. This simulates applying the audit with inaccurate localization data. We sweep the longitudinal standard deviation and adjust the lateral and yaw values linearly with the same ratio used by RTMap for localization-noise simulation~\citep{du2025rtmap}. The nominal noise-trained model was fine-tuned with $(0.85, 0.75, 1.5)$ standard deviations for yaw, lateral, and longitudinal translation, respectively. The heavy-noise model used $(1.7, 1.5, 3.0)$.

The pose-noise curves show that the audit does not collapse as soon as localization is imperfect. At the nominal noise level, which can be seen as consumer-grade accuracy, the same model falls to 86.9 PDMS, while noise-trained variants remain near 88--89 PDMS. Under the largest perturbation, noise training trades clean-pose performance for robustness: the heavy-noise model reaches 84.2 PDMS, compared with 78.5 for the clean-trained model. These scores remain well above DrivoR without cameras, so the diagnostic signal weakens but remains visible under moderate pose noise.

\paragraph{Fewer traversals.} For the \(k\)-sensitivity check, we use the clean MemoryDrivoR checkpoint and keep the retrieval radius fixed at \(r=20\,\mathrm{m}\). Reducing \(k\) limits how much historical evidence is available to the planner at each query. The \(k=0\) point denotes the DrivoR control without cameras.

The \(k\)-sweep shows that the audit also does not require many retrieved memories per query. The main NAVSIMv1 gain appears as soon as one memory is available: \(k=1\) reaches 89.8 PDMS. Increasing \(k\) then gives smaller gains and saturates near 91.1 PDMS for \(k \ge 6\). This suggests that sparse historical coverage can still reveal whether a benchmark score is driven by persistent scene context rather than by observing the current scene.

\subsection{Component Ablation}
\label{app:component-ablation}
\setlength{\intextsep}{5pt}
\begin{wraptable}[10]{R}{0.37\textwidth}
\centering
\captionsetup{font=footnotesize,skip=2pt}
\footnotesize
\begin{threeparttable}
\setlength{\tabcolsep}{2pt}
\caption{\textbf{Ablation of the main components} of our method on NAVSIMv1 \navval{}.}
\label{tab:navsim-v1-ablation}
\begin{scoretabular}{>{\color{gray}\upshape}c>{\color{gray}\upshape}ccccS}
\toprule
$r$ (m) & $k$ & Pose & Resampler & \scorecell{PDMS} & \scorecell{$\Delta$} \\
\midrule
5 & 4 & \xmark & \xmark & \scorecell{82.6} &  \\
5 & 10 & \xmark & \cmark & \scorecell{82.6} & \scorecell{+0.0} \\
20 & 4 & \cmark & \xmark & \scorecell{88.4} & \scorecell{+5.8} \\
20 & 10 & \cmark & \cmark & \scorecell{\textbf{88.9}} & \scorecell{\textbf{+6.3}} \\
\bottomrule
\end{scoretabular}
\end{threeparttable}
\end{wraptable}
We ablate the pose embedder and the resampler, the two main components of MemoryDrivoR, on NAVSIMv1 and report the numbers on the held-out validation set \navval{} in Tab.~\ref{tab:navsim-v1-ablation}. We optimized the retrieval settings for $k$, the maximum number of memories per sample, and the maximum search radius $r$, when no pose information is available or when no resampler is used. Relative-pose conditioning is the dominant factor: removing it keeps PDMS at 82.6 even when the resampler is enabled, while pose-aware retrieval raises performance to 88.4--88.9. The resampler adds a smaller gain on top of pose-aware retrieval.

\subsection{Role of the Resampler}
\label{app:resampler-probe}
To examine how much the resampler filters out dynamic information, we ran an additional probing experiment comparing its output with the original register tokens. As a compression baseline, we also reduced the register tokens using PCA.

For every memory sample, we rasterized a $64\,\mathrm{m}\times64\,\mathrm{m}$ area into a $32\times32$ bird's-eye-view (BEV) grid. For each representation, we trained one decoder for six map classes (centerlines, lane dividers, drivable roadblocks, crosswalks, stop lines, and intersections) and another decoder for three object classes (vehicles, pedestrians, and bicycles). All decoders used 1,024 learned BEV queries, one cross-attention layer, and a two-layer feed-forward block, while the input representations remained frozen. We trained 18 decoders in total (3 representations, 2 target sets, and 3 seeds) using class-weighted binary cross-entropy and AdamW (learning rate $3.0 \times 10^{-4}$, weight decay $10^{-4}$, batch size 32, up to 20 epochs, and early stopping after five epochs). We constructed a probe dataset from NAVSIMv1-Geo by sampling 12,000 training frames and 2,000 validation frames from non-Pittsburgh drives, along with 4,000 test frames from Pittsburgh drives. The memories come from our model trained on the standard NAVSIMv1 split, while the decoders and PCA used this geographical split.

\setlength{\intextsep}{5pt}
\begin{wraptable}[9]{R}{0.39\textwidth}
\vspace{0.5\baselineskip}
\centering
\captionsetup{font=footnotesize,skip=2pt}
\scriptsize
\setlength{\tabcolsep}{2.75pt}
\begin{threeparttable}
\caption{\textbf{Probing memory.}}
\label{tab:resampler-probe}
\begin{scoretabular}{@{}lc@{\hspace{4pt}}*{3}{>{\columncolor{scoregray}[\tabcolsep][\tabcolsep]}Z[table-format=2.1,table-column-width=0.62cm]@{}}>{\columncolor{scoregray}[\tabcolsep][0pt]}Z[table-format=2.1,table-column-width=0.62cm]@{}}
\toprule
\multirow{2}{*}{Representation} & \multirow{2}{*}{Tokens} &
\multicolumn{2}{>{\columncolor{scoregray}[\dimexpr2\tabcolsep\relax][\tabcolsep]}c}{Map} &
\multicolumn{2}{>{\columncolor{scoregray}[\tabcolsep][\tabcolsep]}c}{Agent} \\
 & & {AP} & {IoU} & {AP} & {IoU} \\
\midrule
Raw memory & 64 & 49.7 & 34.1 & 10.9 & 7.8 \\
PCA & 8 & 48.1 & 33.2 & 9.3 & 6.5 \\
Resampled & 8 & 49.1 & 33.9 & 7.9 & 5.5 \\
\bottomrule
\end{scoretabular}
\end{threeparttable}
\end{wraptable}
We report the results in Tab.~\ref{tab:resampler-probe}. Map prediction changes little after the resampler. The average precision (AP) decreases by 0.54 points and the intersection over union (IoU) by 0.17 points. The drop is larger for agents, where AP decreases by 2.94 points and IoU by 2.30 points. PCA removes less agent information while losing more map information. The experiment therefore supports our hypothesis that the resampler keeps static information while reducing information about agents observed during previous drives.

\subsection{Fusing Camera with Memory}

\begin{table}[b]
\centering
\setlength{\tabcolsep}{6pt}
\caption{\textbf{Camera and memory fusion} ablation across benchmarks.}
\label{tab:camera-memory-ablation}
\begin{scoretabular}{@{}lSSSE@{}}
\toprule
\multicolumn{1}{c}{Input} & \scorecell{\scorehead{NAVSIMv1}{PDMS~$\uparrow$}} & \scorecell{\scorehead{NAVSIMv2}{EPDMS~$\uparrow$}} & \scorecell{\scorehead{Bench2Drive}{DS~$\uparrow$}} & \scorecell{\scorehead{Bench2Drive}{SR~$\uparrow$}} \\
\midrule
Mem. & 91.1 & 45.0 & 34.7 & 9.5 \\
Cam. & \textbf{93.7} & 46.7 & 61.0 & 30.5 \\
Cam. + Mem. & \textbf{93.7} & \textbf{47.3} & \textbf{63.2} & \textbf{33.6} \\
\bottomrule
\end{scoretabular}
\end{table}

For completeness, Tab.~\ref{tab:camera-memory-ablation} reports a fusion ablation, where we do not remove the current camera input. Instead, we concatenate retrieved memory tokens with tokens from the evaluated scene and add a learned source flag so that the planner can distinguish memory from current camera input.
During training, we drop all current-scene tokens of a sample with probability $p=0.2$, encouraging the model to use memory when making predictions. If no memory tokens are available for such a sample, the planner skips its cross-attention layers.

The results can be seen as an upper bound for our chosen DrivoR-based planner, where both dynamic and static information are available. The effect of fusion is small across benchmarks, especially compared with the memory gains from the main audits.

\subsection{Qualitative Examples}
\label{app:additional-qualitative-examples}
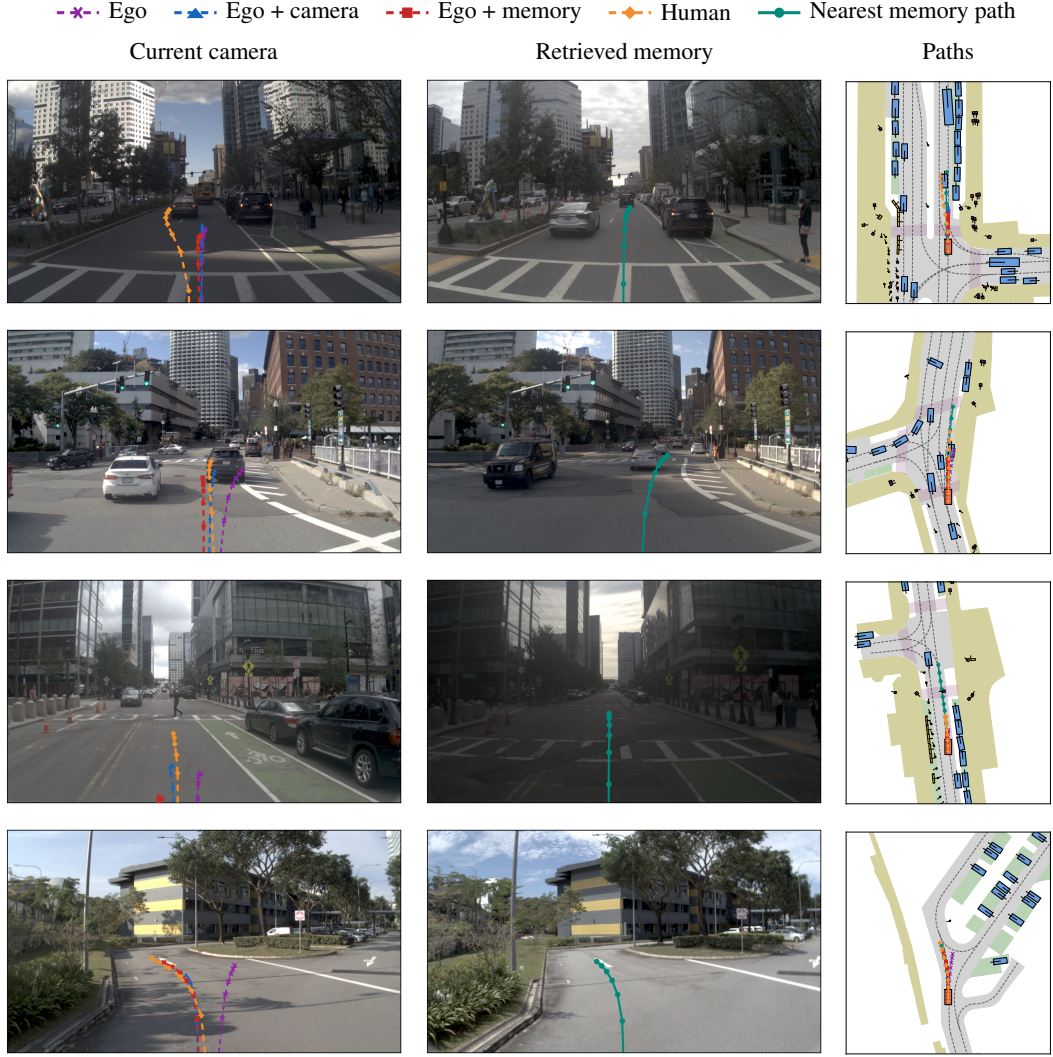
\begin{figure*}[t]
    \centering
    \newcommand{\appendixqualcase}[1]{%
        \makebox[\textwidth][c]{%
            \includegraphics[width=0.386\textwidth]{figures/analysis/dense_traffic_batch3_ego_drivor_camera_latex/#1/scene_image.png}%
            \hspace{0.012\textwidth}%
            \includegraphics[width=0.386\textwidth]{figures/analysis/dense_traffic_batch3_ego_drivor_camera_latex/#1/memory_image.png}%
            \hspace{0.012\textwidth}%
            \includegraphics[width=0.204\textwidth]{figures/analysis/dense_traffic_batch3_ego_drivor_camera_latex/#1/trajectory_plot.pdf}%
        }%
        \par\vspace{0.45em}%
    }
    \makebox[\textwidth][c]{\hspace*{-4.3pt}\definecolor{memonlyred}{HTML}{C62828}
\definecolor{egoonlypurple}{HTML}{8E24AA}
\definecolor{camerablue}{HTML}{1565C0}
\definecolor{memorypathgreen}{HTML}{00897B}
\definecolor{humangtorange}{HTML}{F28E2B}

\begin{tikzpicture}[
    legendline/.style={line width=1.15pt, dash pattern=on 2.8pt off 2.2pt},
    memorylegendline/.style={line width=1.15pt},
    legendtext/.style={anchor=west, font=\small, inner sep=0pt},
    x=1cm,
    y=1cm
]
    \coordinate (ego_start) at (0, 0);
    \draw[legendline, egoonlypurple] (ego_start) -- ++(0.56, 0);
    \draw[egoonlypurple, line width=1.0pt] (ego_start) ++(0.23, -0.07) -- ++(0.10, 0.14);
    \draw[egoonlypurple, line width=1.0pt] (ego_start) ++(0.23, 0.07) -- ++(0.10, -0.14);
    \path (ego_start) ++(0.68, 0) coordinate (ego_text_anchor);
    \node[legendtext] (ego_text) at (ego_text_anchor) {Ego};

    \path (ego_text.east) ++(0.35, 0) coordinate (camera_start);
    \draw[legendline, camerablue] (camera_start) -- ++(0.56, 0);
    \path (camera_start) ++(0.28, 0.055) coordinate (camera_marker_top);
    \path (camera_start) ++(0.18, -0.065) coordinate (camera_marker_left);
    \path (camera_start) ++(0.38, -0.065) coordinate (camera_marker_right);
    \filldraw[camerablue] (camera_marker_top) -- (camera_marker_left) -- (camera_marker_right) -- cycle;
    \path (camera_start) ++(0.68, 0) coordinate (camera_text_anchor);
    \node[legendtext] (camera_text) at (camera_text_anchor) {Ego + camera};

    \path (camera_text.east) ++(0.35, 0) coordinate (mem_start);
    \draw[legendline, memonlyred] (mem_start) -- ++(0.56, 0);
    \path (mem_start) ++(0.28, 0) coordinate (mem_marker);
    \node[draw=memonlyred, fill=memonlyred, minimum size=3.5pt, inner sep=0pt] at (mem_marker) {};
    \path (mem_start) ++(0.68, 0) coordinate (mem_text_anchor);
    \node[legendtext] (mem_text) at (mem_text_anchor) {Ego + memory};

    \path (mem_text.east) ++(0.35, 0) coordinate (gt_start);
    \draw[legendline, humangtorange] (gt_start) -- ++(0.56, 0);
    \path (gt_start) ++(0.28, 0) coordinate (gt_marker);
    \node[diamond, draw=humangtorange, fill=humangtorange, minimum size=4.5pt, inner sep=0pt] at (gt_marker) {};
    \path (gt_start) ++(0.68, 0) coordinate (gt_text_anchor);
    \node[legendtext] (gt_text) at (gt_text_anchor) {Human};

    \path (gt_text.east) ++(0.35, 0) coordinate (memory_start);
    \draw[memorylegendline, memorypathgreen] (memory_start) -- ++(0.56, 0);
    \path (memory_start) ++(0.28, 0) coordinate (memory_marker);
    \node[circle, fill=memorypathgreen, minimum size=3.8pt, inner sep=0pt] at (memory_marker) {};
    \path (memory_start) ++(0.68, 0) coordinate (memory_text_anchor);
    \node[legendtext] at (memory_text_anchor) {Nearest memory path};
\end{tikzpicture}}
    \par\vspace{0.25em}
    \makebox[\textwidth][c]{%
        \begin{minipage}[t]{0.386\textwidth}
            \centering
            \small Current camera
        \end{minipage}%
        \hspace{0.012\textwidth}%
        \begin{minipage}[t]{0.386\textwidth}
            \centering
            \small Retrieved memory
        \end{minipage}%
        \hspace{0.012\textwidth}%
        \begin{minipage}[t]{0.204\textwidth}
            \centering
            \small Paths
        \end{minipage}%
    }
    \par\vspace{0.25em}
    \appendixqualcase{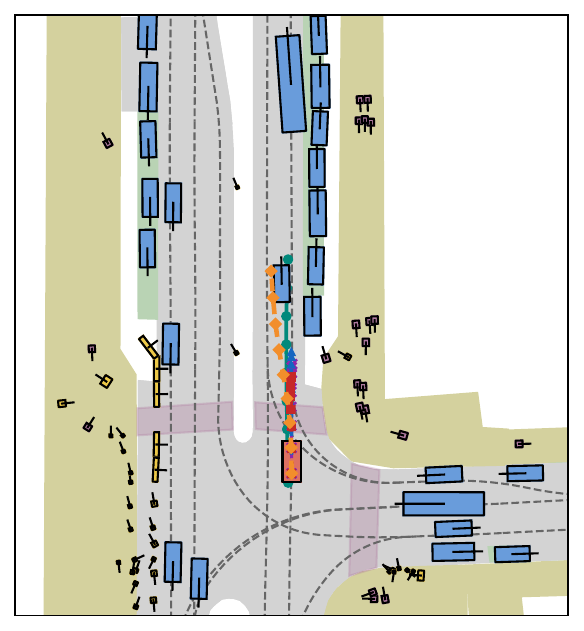}
    \appendixqualcase{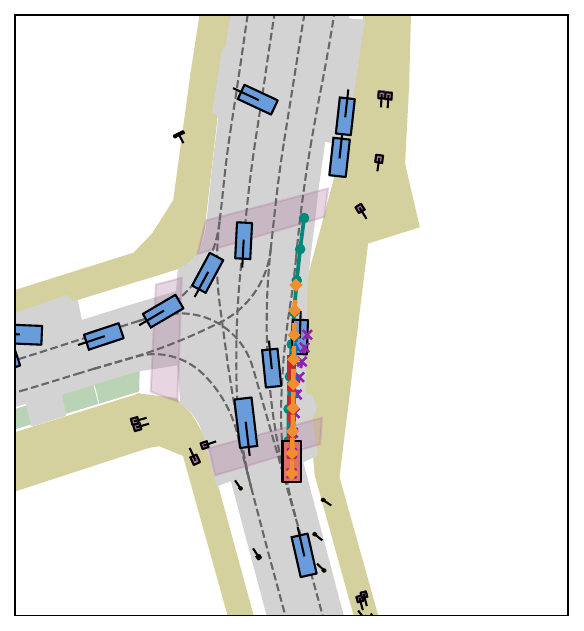}
    \appendixqualcase{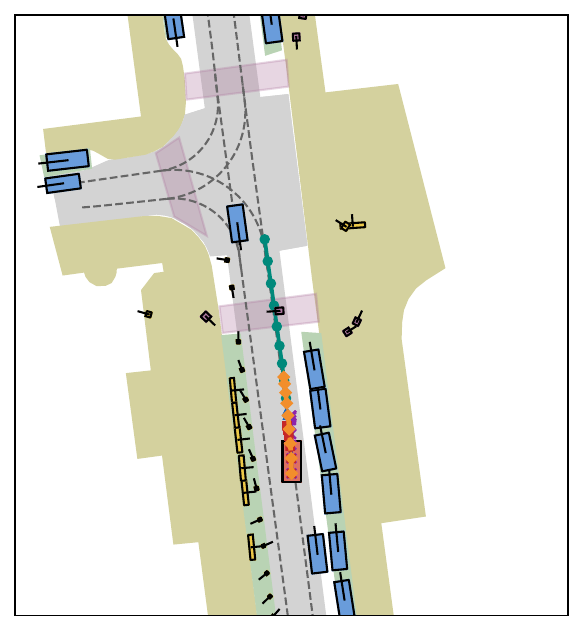}
    \appendixqualcase{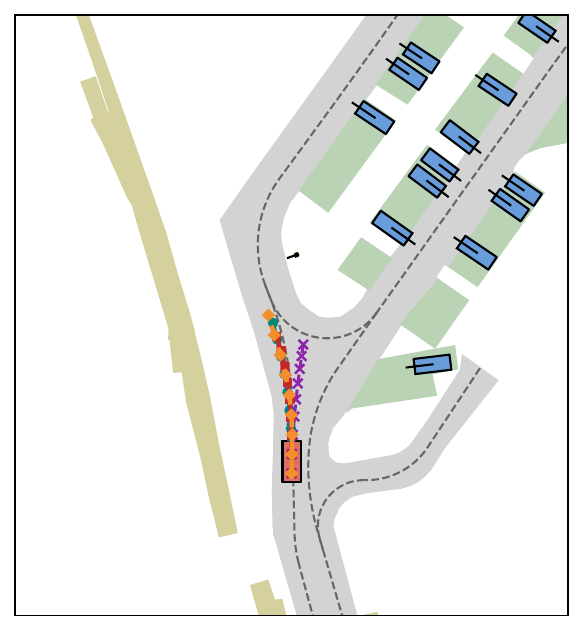}
    \caption{\textbf{Additional examples} comparing DrivoR inputs on NAVSIMv1. Each row follows the format of Fig.~\ref{fig:mem_beats_ego_example}: the left panel shows the evaluated scene, the middle panel shows the closest retrieved memory, and the right panel overlays ego-only, camera-based, memory-based, and human-reference trajectories over the tested horizon. Compared with the camera-based model, MemoryDrivoR tends to drive more conservatively in locations where dense traffic is expected, but can make more progress in less crowded locations.}

    \label{fig:appendix-dense-traffic-cases}
\end{figure*}

Fig.~\ref{fig:appendix-dense-traffic-cases} shows additional NAVSIMv1 cases comparing MemoryDrivoR with DrivoR and its control without cameras. The examples illustrate that MemoryDrivoR often slows down where dense traffic is likely, but can be more progressive when the retrieved scene suggests less congestion.

\section{Additional Implementation Details}

\subsection{Memory Retrieval}
\label{app:memory-retrieval-statistics}
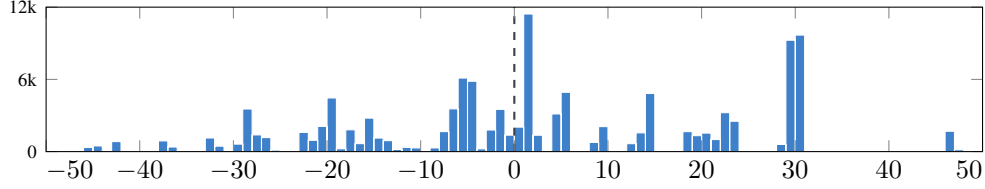
\begin{figure}[t]
  \centering
  \pgfplotstableread{%
center count
-60.5 0
-59.5 0
-58.5 0
-57.5 0
-56.5 0
-55.5 0
-54.5 0
-53.5 0
-52.5 0
-51.5 210
-50.5 432
-49.5 0
-48.5 0
-47.5 152
-46.5 1691
-45.5 0
-44.5 0
-43.5 0
-42.5 0
-41.5 0
-40.5 0
-39.5 0
-38.5 0
-37.5 0
-36.5 0
-35.5 0
-34.5 0
-33.5 0
-32.5 0
-31.5 37
-30.5 9681
-29.5 9253
-28.5 594
-27.5 0
-26.5 0
-25.5 0
-24.5 0
-23.5 2522
-22.5 3234
-21.5 1001
-20.5 1538
-19.5 1324
-18.5 1658
-17.5 0
-16.5 0
-15.5 0
-14.5 4826
-13.5 1557
-12.5 659
-11.5 0
-10.5 0
-9.5 2075
-8.5 762
-7.5 0
-6.5 0
-5.5 4928
-4.5 3128
-3.5 0
-2.5 1361
-1.5 11431
-0.5 2032
0.5 1365
1.5 3510
2.5 1795
3.5 228
4.5 5844
5.5 6111
6.5 3550
7.5 1661
8.5 306
9.5 0
10.5 306
11.5 352
12.5 174
13.5 910
14.5 1126
15.5 2782
16.5 661
17.5 1811
18.5 231
19.5 4456
20.5 2095
21.5 940
22.5 1593
23.5 0
24.5 0
25.5 123
26.5 1164
27.5 1387
28.5 3539
29.5 622
30.5 0
31.5 452
32.5 1130
33.5 0
34.5 0
35.5 3
36.5 380
37.5 889
38.5 0
39.5 0
40.5 0
41.5 0
42.5 832
43.5 1
44.5 470
45.5 345
46.5 0
47.5 0
48.5 0
49.5 0
50.5 49
51.5 1761
52.5 218
53.5 0
54.5 0
55.5 0
56.5 0
57.5 0
58.5 0
59.5 94
60.5 108
}\timedeltatable

\begin{tikzpicture}
\begin{axis}[
    width=\textwidth,
    height=0.25\linewidth,
    xmin=-50,
    xmax=50,
    ymin=0,
    ymax=12000,
    ybar,
    bar width=1,
    enlarge x limits=false,
    grid=none,
    major grid style={draw=black!18},
    minor grid style={draw=black!8},
    tick align=inside,
    axis on top=true,
    axis line style={draw=black},
    xtick={-50,-40,-30,-20,-10,0,10,20,30,40,50},
    xticklabels={\makebox[0pt][l]{$-50$},$-40$,$-30$,$-20$,$-10$,$0$,$10$,$20$,$30$,$40$,\makebox[0pt][r]{$50$}},
    ytick={0,6000,12000},
    yticklabels={0,6k,12k},
    yticklabel style={font=\scriptsize},
    scaled y ticks=false,
]
\addplot[
    draw=white,
    fill=plotblue,
    fill opacity=0.82,
    x filter/.code={\pgfmathparse{-#1}},
] table[x=center, y=count] {\timedeltatable};
\draw[dashed, thick, black!75] (axis cs:0,0) -- (axis cs:0,12000);
\end{axis}
\end{tikzpicture}
  \caption{\textbf{Distribution of relative timestamps in days} for all retrieved memories in NAVSIMv1 \navtest{}. Negative values indicate memories from future training samples.}
  \label{fig:memory-time-delta}
\end{figure}

For both NAVSIM versions, we construct the memory bank from all samples in the standard training split, \navtrain{}. The complete bank occupies only \SI{6.3}{GB} across all four cities combined, which is smaller than the checkpoints of tested vision-language-action models such as AutoVLA~\citep{zhou2025autovla} and ReCogDrive~\citep{li2026recogdrive}. Most retrieved memories come from drives recorded several weeks before or after the evaluated sample, as shown in Fig.~\ref{fig:memory-time-delta}.

\begin{figure}[t]
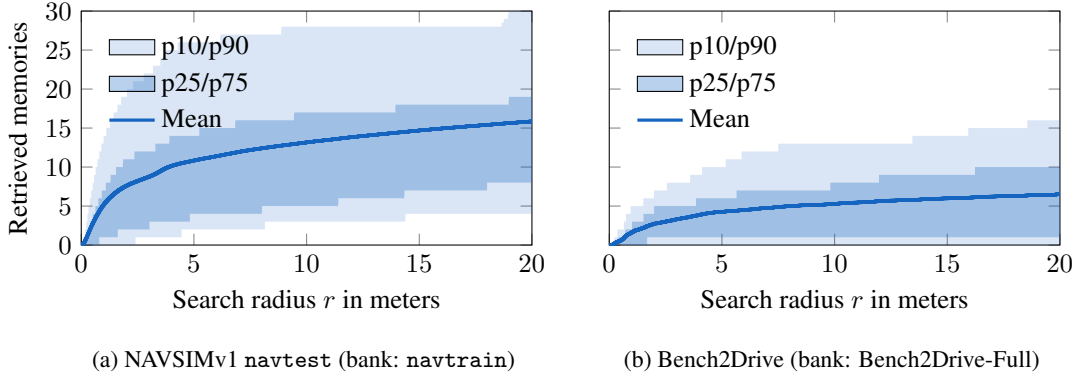

  \centering
  \makebox[\textwidth][c]{%
    \begin{subfigure}[t]{0.54\textwidth}
      \centering
      \input{figures/avg_memory_count_vs_distance_threshold.tikz.tex}
      \begin{minipage}[t][2.8ex][t]{\linewidth}
        \centering
        \captionsetup{margin={0.095\linewidth,0pt},skip=1pt}
        \caption{NAVSIMv1 \navtest{} (bank: \navtrain{})}
        \label{fig:retrieval-count-threshold-ns1}
      \end{minipage}
    \end{subfigure}\hspace{-0.04\textwidth}%
    \begin{subfigure}[t]{0.54\textwidth}
      \centering
      \input{figures/avg_memory_count_vs_distance_threshold_b2d.tikz.tex}
      \begin{minipage}[t][2.8ex][t]{\linewidth}
        \centering
        \captionsetup{margin={0.095\linewidth,0pt},skip=1pt}
        \caption{Bench2Drive (bank: Bench2Drive-Full)}
        \label{fig:retrieval-count-threshold-b2d}
      \end{minipage}
    \end{subfigure}%
  }
  \caption{\textbf{Memory retrieval.} Average number of memories retrieved per test sample within a search radius $r$. Bands show the 10th/90th and 25th/75th percentiles. Bank denotes the data split used for the memory bank. Bench2Drive numbers can vary because evaluation is closed loop.}
  \label{fig:retrieval-count-threshold}
\end{figure}

The standard Bench2Drive training set has sparser repeated-location coverage than NAVSIM, so we build the memory bank from the Bench2Drive-Full training dataset containing 10 times more drives. Because the dataset is sampled at \SI{10}{Hz}, we subsample before storage and keep a memory only after the ego vehicle has moved at least \SI{1.5}{\meter} or after \SI{5}{\second} has elapsed.
We also exclude extreme weather and low-light conditions by filtering the bank to Bench2Drive training scenes with weather IDs $\{0\text{--}3,5\text{--}7,15,18,26\}$. After these filters, the Bench2Drive bank occupies \SI{21.2}{GB} across all \(12\) towns. Fig.~\ref{fig:retrieval-count-threshold} reports the average number of retrieved memories for both benchmarks.

\subsection{DrivoR Implementation for Bench2Drive}
\label{app:bench2drive-drivor}
For training with Bench2Drive, we keep DrivoR's proposal-based planner and loss structure, but replace the NAVSIM simulation-based target scores with simpler non-reactive geometric targets for the same score heads. Unlike NAVSIM, these targets do not include driving-direction compliance.

As with NAVSIM, the model predicts 64 candidate trajectories, each represented by 6 waypoints over a \SI{3}{\second} horizon. The vision-based model also uses four RGB cameras: front, front-left, front-right, and rear. The trajectory loss is a best-of-64 L1 imitation loss on the predicted trajectories. Score heads are trained with binary cross-entropy losses for no-collision, drivable-area compliance, time-to-collision, ego progress, and comfort.

For MemoryDrivoR and the DrivoR baseline without camera, the internal trajectory score often chose stationary trajectories, i.e., the ego vehicle never started to drive. We therefore use progress-only proposal selection, setting the subscore weights of all other components to zero. This changes only the model's trajectory selection, not the closed-loop Bench2Drive evaluation metric. It improves the driving score from 0.5 to 11.6 for ego-status DrivoR and from 26.6 to 34.7 for MemoryDrivoR. For camera-based DrivoR, this did not improve performance, so we kept the original subscore weights.

We train the DrivoR base model for 20 epochs with AdamW, a global batch size of 64, and a learning rate of \(10^{-4}\). For the memory-based fine-tuning, we use the same settings. For full reproducibility, we refer the reader to our code.

\subsection{HD-Map Baseline}
\label{app:hd-map}
For the HD-map baseline, we represent the static map from nuPlan~\citep{karnchanachari2024nuplan} as a set of vector instances around the ego vehicle. Lane and lane-connector objects are decomposed into centerline, left-boundary, and right-boundary polylines. We also include static polygonal elements such as stop lines, crosswalks, intersections, roadblocks, roadblock connectors, walkways, and car-park areas. Each map instance is transformed into the ego frame and uniformly sampled to a fixed number of points.

Each instance is encoded with its semantic element type, geometry type, and available static attributes. The attributes include route membership, whether the element is traffic-light controlled, stop-line subtype, and speed limit when available. For lane and lane-connector elements, we additionally encode local direction from the baseline heading. We do not use dynamic traffic-light state, surrounding actors, camera features, or any other current-scene observation beyond the ego-status input.

The encoder follows the main idea of PriorDrive's Unified Vector Encoder~\citep{zeng2026priordrive}, which treats vectorized map elements as ordered point sequences and uses attention to capture both within-instance structure and relationships across instances. In our adaptation, coordinates and directions are embedded with random Fourier features~\citep{tancik2020fourier}. We add learned embeddings for the geometry type, semantic map-element type, route-membership flag, traffic-light-control flag, and stop-line subtype, and project the scalar speed limit when available. A BERT encoder~\citep{devlin2019bert} first processes the points within each map instance. A second BERT encoder then processes the resulting instance tokens across the local map. The final map tokens are projected to DrivoR's planner dimension $c$ and used as the planner's keys $K$ and values $V$ in place of camera-derived scene tokens.

Tab.~\ref{tab:appendix-hdmap-navsim-v1} and Tab.~\ref{tab:appendix-hdmap-navsim-v2} expand the aggregate HD-map comparison in Tab.~\ref{tab:hdmap-comparison} with the full NAVSIMv1 and NAVSIMv2 component scores.

\begin{table}[t]
\centering
\setlength{\tabcolsep}{3pt}
\begin{threeparttable}
\caption{\textbf{Full NAVSIMv1 HD-map results} compared with other DrivoR inputs.}
\label{tab:appendix-hdmap-navsim-v1}
\begin{scoretabular}{@{}lcccccE@{}}
\toprule
\multicolumn{1}{c}{Input} & NC & DAC & TTC & C & EP & \scorecell{PDMS~$\uparrow$} \\
\midrule
\refrowcell{Ego + cameras} & \refrowcell{99.0} & \refrowcell{98.9} & \refrowcell{96.7} & \refrowcell{100.0} & \refrowcell{90.0} & \scorecell{\refrowcell{93.7}} \\
\midrule
Ego & 96.1 & 85.6 & 91.1 & 100.0 & 62.8 & \scorecell{72.9} \\
Ego + HD map & 97.9 & 96.3 & 94.4 & 100.0 & 79.2 & \scorecell{86.6} \\
Ego + memory & 98.3 & 98.5 & 95.3 & 100.0 & 86.0 & \scorecell{\textbf{91.1}} \\
\bottomrule
\end{scoretabular}
\end{threeparttable}
\end{table}

\begin{table}[t]
\centering
\setlength{\tabcolsep}{2pt}
\begin{threeparttable}
\caption{\textbf{Full NAVSIMv2 HD-map results} compared with other DrivoR inputs.}
\label{tab:appendix-hdmap-navsim-v2}
\begin{scoretabular}{@{}>{\raggedright\arraybackslash}p{1.4cm}ccccccccccE@{}}
\toprule
\multicolumn{1}{c}{\makebox[1.4cm]{Input}} & Stage & NC & DAC & DDC & TLC & EP & TTC & LK & HC & EC & \scorecell{EPDMS~$\uparrow$} \\
\midrule
\multirow{2}{=}{\refrowcell{Ego + cameras}} & \refrowcell{S1} & \refrowcell{99.3} & \refrowcell{95.8} & \refrowcell{99.3} & \refrowcell{99.8} & \refrowcell{73.5} & \refrowcell{99.3} & \refrowcell{94.2} & \refrowcell{97.6} & \refrowcell{70.7} & \\
 & \refrowcell{S2} & \refrowcell{89.8} & \refrowcell{86.4} & \refrowcell{90.8} & \refrowcell{98.3} & \refrowcell{71.3} & \refrowcell{88.2} & \refrowcell{51.8} & \refrowcell{99.0} & \refrowcell{76.8} & \scorecell{\multirow{-2}{*}{\refrowcell{46.7}}} \\
\midrule
\multirow{2}{=}{Ego} & S1 & 95.4 & 72.2 & 93.7 & 100.0 & 67.4 & 94.4 & 83.8 & 97.6 & 68.4 & \\
 & S2 & 88.2 & 71.7 & 87.8 & 99.3 & 56.1 & 86.1 & 48.1 & 99.0 & 80.9 & \scorecell{\multirow{-2}{*}{28.3}} \\
\multirow{2}{=}{Ego + \\HD map} & S1 & 97.8 & 90.9 & 97.3 & 100.0 & 71.8 & 97.1 & 90.4 & 97.6 & 76.0 & \\
 & S2 & 88.3 & 84.8 & 92.1 & 99.3 & 59.5 & 87.6 & 51.4 & 98.0 & 74.4 & \scorecell{\multirow{-2}{*}{40.8}} \\
\multirow{2}{=}{Ego + memory} & S1 & 98.2 & 95.8 & 99.0 & 100.0 & 72.0 & 98.0 & 92.0 & 97.6 & 69.3 & \\
 & S2 & 88.2 & 89.1 & 92.2 & 98.7 & 58.4 & 87.6 & 52.0 & 98.3 & 75.2 & \scorecell{\multirow{-2}{*}{\textbf{45.0}}} \\
\bottomrule
\end{scoretabular}
\end{threeparttable}
\end{table}

\subsection{Compute Resources}
\label{appendix:compute-resources}

Tab.~\ref{tab:compute-resources} summarizes the completed runs used for memory-bank construction, training, and evaluation. NAVSIM evaluation is offline, whereas Bench2Drive, HUGSIM-nuScenes, and RealEngine use closed-loop simulation. The NAVSIM memory bank adds \SI{6.3}{GB} of storage across the four cities, and the filtered Bench2Drive memory bank adds \SI{21.2}{GB} across all \(12\) towns, excluding the original benchmark datasets. The HUGSIM-nuScenes memory bank adds \SI{2.2}{GB} for \(34{,}149\) nuScenes trainval keyframes. RealEngine reuses the NAVSIMv1 memory bank. The NAVSIMv1 MemoryDrivoR run uses the original DrivoR checkpoint, and the HUGSIM-nuScenes and RealEngine evaluations use frozen NAVSIMv1 checkpoints without simulator-specific training.

\begin{table}[t]
\centering
\footnotesize
\setlength{\tabcolsep}{3pt}
\begin{threeparttable}
\caption{\textbf{Compute resources} for memory construction, training, and evaluation.}
\label{tab:compute-resources}
\begin{scoretabular}{@{}llN{2.0}N{1.0}lN{2.0}N{2.1}N{3.1}@{}}
\toprule
\multicolumn{1}{c}{Benchmark} & \multicolumn{1}{c}{Run} & \multicolumn{1}{c}{CPUs} & \multicolumn{1}{c}{GPUs} & \multicolumn{1}{c}{GPU type} & \multicolumn{1}{c}{Epochs} & \multicolumn{1}{c}{\scorehead{Wall}{time (h)}} & \multicolumn{1}{c}{\scorehead{Peak mem.}{(GB)}} \\
\midrule
\multicolumn{8}{@{}l@{}}{\textit{Memory bank construction}} \\
NAVSIMv1 & Memory bank & 5 & 2 & RTX 6000 Ada & \multicolumn{1}{c}{--} & 3.5 & 39.8 \\
NAVSIMv2 & Memory bank & 5 & 2 & RTX 3090 & \multicolumn{1}{c}{--} & 4.5 & 40.3 \\
Bench2Drive & Memory bank & 20 & 2 & H200 MIG & \multicolumn{1}{c}{--} & 1.9 & 105.8 \\
HUGSIM-nuScenes & Memory bank & 9 & 4 & RTX A6000 & \multicolumn{1}{c}{--} & 0.6 & 32.9 \\
\midrule
\multicolumn{8}{@{}l@{}}{\textit{Training}} \\
NAVSIMv2 & Reproduced DrivoR & 20 & 4 & RTX A6000 & 10 & 56.1 & 114.6 \\
NAVSIMv1 & MemoryDrivoR & 20 & 4 & RTX 6000 Ada & 5 & 54.0 & 132.8 \\
NAVSIMv2 & MemoryDrivoR & 20 & 4 & RTX A6000 & 5 & 80.7 & 133.0 \\
Bench2Drive & Reimplemented DrivoR & 20 & 2 & H200 MIG & 20 & 18.3 & 47.8 \\
Bench2Drive & MemoryDrivoR & 20 & 2 & H200 MIG & 20 & 15.2 & 91.5 \\
\midrule
\multicolumn{8}{@{}l@{}}{\textit{Evaluation}} \\
NAVSIMv1 & MemoryDrivoR & 3 & 1 & RTX 3090 & \multicolumn{1}{c}{--} & 1.4 & 18.3 \\
NAVSIMv2 & MemoryDrivoR & 3 & 1 & RTX A5000 & \multicolumn{1}{c}{--} & 1.3 & 37.2 \\
Bench2Drive & DrivoR & 4 & 1 & A100 80GB & \multicolumn{1}{c}{--} & 45.1 & 9.4 \\
Bench2Drive & MemoryDrivoR & 4 & 1 & A100 80GB & \multicolumn{1}{c}{--} & 42.2 & 30.3 \\
RealEngine & DrivoR & 8 & 1 & RTX 6000 Ada & \multicolumn{1}{c}{--} & 0.3 & 12.0 \\
RealEngine & MemoryDrivoR & 4 & 1 & RTX A6000 & \multicolumn{1}{c}{--} & 0.2 & 8.0 \\
HUGSIM-nuScenes & DrivoR & 1 & 1 & RTX 6000 Ada & \multicolumn{1}{c}{--} & 1.2 & 3.8 \\
HUGSIM-nuScenes & MemoryDrivoR & 9 & 1 & RTX 6000 Ada & \multicolumn{1}{c}{--} & 2.7 & 6.5 \\
\bottomrule
\end{scoretabular}
\end{threeparttable}
\end{table}

For the Bench2Drive evaluation rows, CPUs and GPUs are per route job. The listed wall time is the final completed evaluation summed over \(220\) route jobs plus aggregation. Each HUGSIM-nuScenes evaluation row corresponds to one full-suite job covering all \(87\) scenarios plus aggregation. The simulator and planner processes share the listed GPU. For RealEngine, the values cover all evaluation settings within one job. The table excludes preliminary, failed, video-rendering, and exploratory runs, so the full research project used more compute than the completed runs reported here.

\subsection{Existing Asset Licenses}
\label{app:asset-licenses}

We use existing assets for non-commercial academic evaluation and do not redistribute raw benchmark or simulator data. We build on DrivoR and use its released NAVSIM checkpoints to initialize MemoryDrivoR. The license for the DrivoR code and checkpoints is Apache License 2.0~\citep{kirby2026drivor,drivorRepository}.
NAVSIM code and benchmark assets are released under the Apache License 2.0. Its OpenScene v1.1 data inherit OpenScene and nuPlan terms: CC BY-NC-SA 4.0 plus the nuPlan Dataset License Agreement for Non-Commercial Use. The nuPlan devkit is Apache License 2.0~\citep{navsimRepository,opensceneRepository,nuplanDevkit}.

For Bench2Drive V0.0.3 and Bench2Drive-Full, its official repository states CC BY-NC-ND 4.0, while the Hugging Face dataset cards currently list Apache License 2.0. We document both and follow the more restrictive repository notice~\citep{bench2driveRepository,bench2driveHFDataset,bench2driveFullHFDataset}. Bench2Drive runs on CARLA 0.9.15 and AdditionalMaps\_0.9.15. CARLA code is under the MIT License, CARLA assets are under CC-BY, and Unreal Engine 4 follows its own terms~\citep{dosovitskiy2017carla,carla0915Repository,carla0915Release}.

RealEngine code is released under the MIT License~\citep{realengineRepository}. Its bundled Street Gaussians \cite{streetGaussiansLicense} camera-rendering code is restricted to educational, research, and non-profit use. HUGSIM code~\citep{hugsimRepository} and its Hugging Face distribution~\citep{hugsimHFDataset} of scenes, scenarios, and reconstructed 3DRealCar models are released under the MIT License, while the underlying 3DRealCar dataset is released under the Apache License 2.0~\citep{threeDRealCarRepository}. The HUGSIM-nuScenes scene reconstructions derive from nuScenes data and therefore remain subject to the linked CC BY-NC-SA 4.0 license and the additional nuScenes Dataset Terms~\citep{nuscenesTerms}.

\section{Societal Impacts}
\label{app:societal-impacts}

This work introduces a method for auditing benchmarks, not a deployed driving system. Its main positive effect is methodological: it can help benchmark designers, reviewers, and practitioners identify when high autonomous-driving benchmark scores can be obtained from ego status and persistent scene context without much reliance on the current traffic scene. Making this failure mode visible can reduce overconfidence in leaderboard progress.

The result can also be misused or misread. A practitioner could treat the NAVSIM findings as evidence that cameras or understanding of the current scene are less important for real-world driving, or could optimize a model for benchmark score by exploiting repeated locations and static regularities. Such uses would be unsafe. Our experiments support only a benchmark-level claim, and the Bench2Drive contrast shows that the effect is benchmark dependent.

\clearpage

\end{document}